\documentclass{article}

\makeatletter
\renewcommand\paragraph{\@startsection{paragraph}{4}{\z@}
	{0.7ex \@plus.2ex \@minus.2ex}
	{-1em}
	{\normalfont\normalsize\bfseries\maybe@addperiod}}
\newcommand{\maybe@addperiod}[1]{#1\@addpunct{.}}
\makeatother

\usepackage{iclr_template/iclr2026_conference,times}

\usepackage[utf8]{inputenc}
\usepackage[T1]{fontenc}

\definecolor{newlightblue}{rgb}{0.21,0.49,0.74}
\usepackage[pagebackref,breaklinks,colorlinks, urlcolor={newlightblue}, citecolor={newlightblue}]{hyperref}
\usepackage{url}
\usepackage{booktabs}
\usepackage{amsfonts}
\usepackage{nicefrac}
\usepackage{microtype}
\usepackage[table]{xcolor}
\usepackage{enumitem}
\usepackage{multirow}
\usepackage[font=small]{caption}
\usepackage{pifont}
\usepackage{graphicx}
\usepackage{tabularx}
\usepackage{amsmath}
\usepackage{wrapfig}
\usepackage{arydshln}

\usepackage{float}

\usepackage{mathtools}

\DeclarePairedDelimiter\floor{\lfloor}{\rfloor}

\usepackage[dvipsnames]{xcolor}

\newcommand{\mypar}[1]{\vspace{-2.5mm}\paragraph{#1}}
\newcommand{\mysection}[1]{\vspace{-2mm}\section{#1}\vspace{-1mm}}
\newcommand{\mysubsection}[1]{\vspace{-1.5mm}\subsection{#1}\vspace{-.75mm}}

\newcommand{\xhdr}[1]{\vspace{4pt}\noindent\textbf{#1}}

\newcommand{\cmark}{\ding{51}}

\definecolor{graycolor}{gray}{.9}
\newcommand{\grayback}[1]{\cellcolor{graycolor}{#1}}

\title{Point Prompting: Counterfactual \\ Tracking with Video Diffusion Models}

\author{Ayush Shrivastava\textsuperscript{1}\thanks{Equal contribution.}\, , Sanyam Mehta\textsuperscript{1}\footnotemark[1]\ , Daniel Geng\textsuperscript{1}, Andrew Owens\textsuperscript{1,2}
\\\textsuperscript{1}University of Michigan, \textsuperscript{2}Cornell University\\ {\normalsize \texttt{\url{https://point-prompting.github.io}}}
\vspace{-5mm}
}

\iclrfinalcopy

\begin{document}
\maketitle

\begin{abstract}
Trackers and video generators solve closely related problems: the former analyze motion, while the latter synthesize it. We show that this connection enables pretrained video diffusion models to perform zero-shot point tracking by simply prompting them to visually mark points as they move over time. We place a distinctively colored marker at the query point, then regenerate the rest of the video from an intermediate noise level. This propagates the marker across frames, tracing the point's trajectory. To ensure that the marker remains visible in this counterfactual generation, despite such markers being unlikely in natural videos, we use the unedited initial frame as a negative prompt. Through experiments with multiple image-conditioned video diffusion models, we find that these ``emergent'' tracks outperform those of prior zero-shot methods and persist through occlusions, often obtaining performance that is competitive with specialized self-supervised models.
Finally, we show that trajectories produced by pretrained generators can be distilled into a fast tracker with similar performance, serving as effective supervision for a tracking model.
\end{abstract}

\vspace{-1mm}
\mysection{Introduction}
\vspace{-1mm}
Recent generative models have shown the remarkable ability to produce temporally consistent videos. The objects within them persist across frames, through occlusion, and despite variations in camera pose and lighting. These capabilities are closely related to the {\em visual tracking} problem. While generation deals with producing videos that contain temporally persistent objects, tracking deals with analyzing such videos to estimate motion. A variety of methods have exploited the connections between these two problems, such as by using trackers to supervise or control video generators~\citep{Chefer2025VideoJAMJA,burgert2025go,geng2025motion,hao2018controllable,ardino2021click} and to evaluate the temporal consistency of generated videos  by measuring how ``trackable'' they are~\citep{allen2025direct,Lai-ECCV-2018,ceylan2023pix2video,tokenflow2023}.

\begin{figure*}[t]
    \centering
    \includegraphics[width=\textwidth]{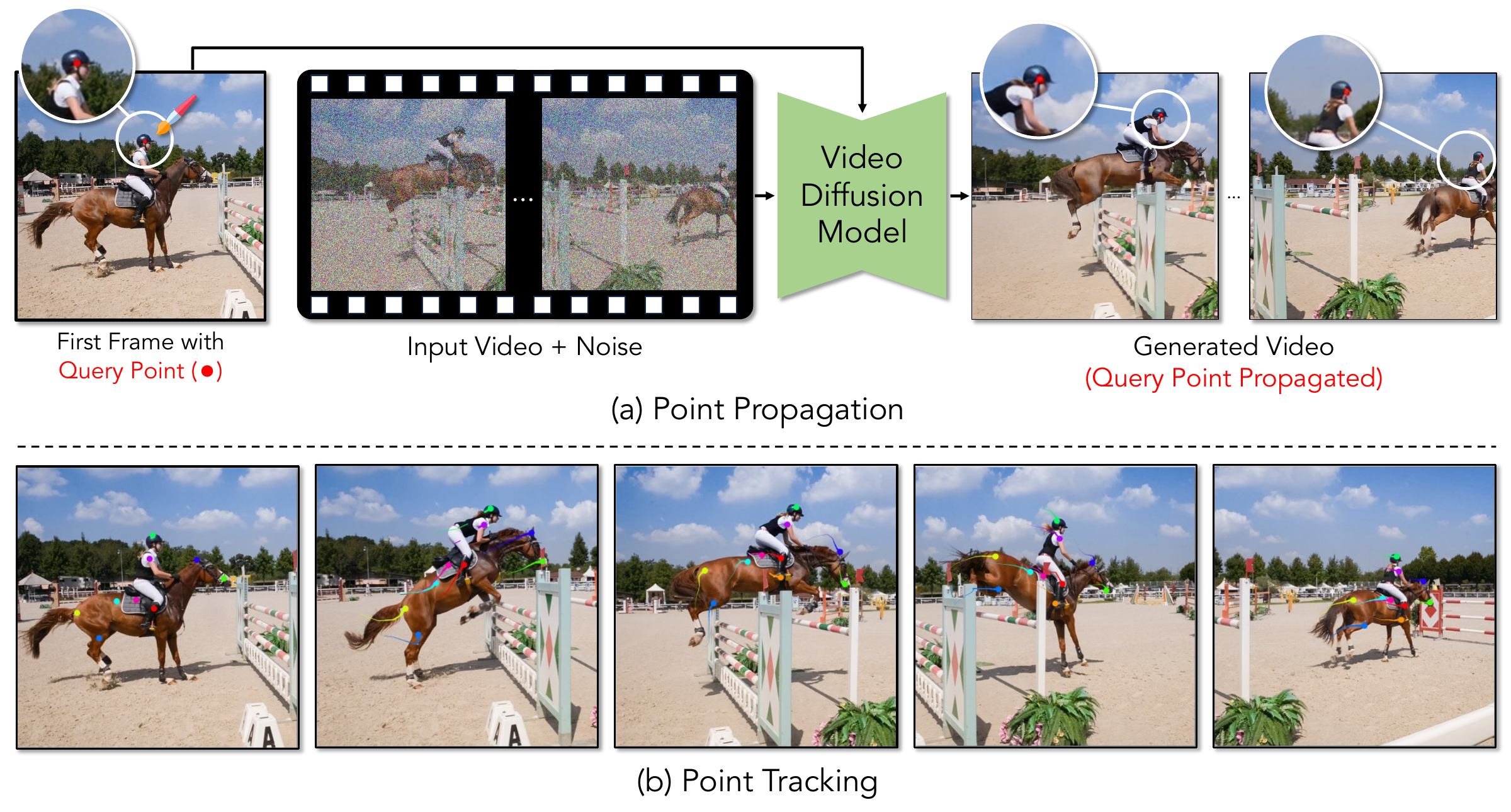}\vspace{-3mm}
    \caption{{\bf Prompting a diffusion model for tracking}. (a) We use an off-the-shelf video diffusion model to perform point tracking. We add a small, distinctive marking---a red dot---to the first frame of an input video, then ask the diffusion model to regenerate the rest of the video using SDEdit~\citep{meng2021sdedit}, which propagates the marking to subsequent frames. (b) We then track the motion of this marking over time. This motion corresponds to the trajectory of the underlying physical point. The model successfully tracks through occlusion. Please see the webpage for more results: \href{https://point-prompting.github.io}{https://point-prompting.github.io}.}
    \label{fig:teaser}
    \vspace{-5mm}
\end{figure*}

In this paper, we ask whether tracking capabilities {\em emerge automatically} in video diffusion models, as a consequence of the close connection between the two problems. Unlike high-level understanding tasks that are naturally described by captions, like object recognition, tracking cannot easily be induced by text prompting. To elicit these capabilities from a video generator, we propose a novel approach to {\em counterfactual modeling} that allows us to directly obtain high-quality point tracks ``zero shot'' from pretrained image-conditioned video diffusion models. We simply mark the position of the query point in the initial video frame using a distinctively colored dot (Fig.~\ref{fig:teaser}), then propagate it to future video frames by regenerating the video using SDEdit~\citep{meng2021sdedit}. After generation, the query point's position can be estimated in each frame by basic image processing.

\looseness=-1
In counterfactual modeling~\citep{bear2023unifying}, one carefully perturbs the input variables, then analyzes how the generation changes in response. Yet large generative models have strong priors that sometimes conflict with this goal. The marker in Fig.~\ref{fig:teaser}, for example, may be unnatural in some environments, and so samples from a generative model may ignore it.
We use a simple but effective method to address this issue: when sampling from the model, we use the unmodified initial input frame as a negative prompt for the diffusion model, thereby guiding the model toward samples that contain the marker.

Our approach is closely related to (and takes inspiration from) a recent line of work that applies counterfactual modeling to self-supervised motion estimation~\citep{bear2023unifying,venkatesh2023understanding}. These methods train a future prediction model, then measure how the predicted future changes when a given point is perturbed in the initial frame, indicating its motion. This requires training a special-purpose model (based on masked autoencoders) that is designed specifically with this downstream use case in mind, and requires training auxiliary models to obtain high performance. By contrast, we show that {\em off-the-shelf} video diffusion models can track points. In this way, our work is closely related to zero-shot emergent correspondence methods~\citep{tang2023emergent, zhang2023tale}. However, these methods treat the pretrained models as representation learners: they extract their internal features and use them to match pairs of images. Instead, we prompt a video diffusion model to visually mark point trajectories. 

Our results suggest that video diffusion models are capable of tracking points through video via counterfactual modeling, without need for additional training. Through experiments on the TAP-Vid~\citep{doersch2022tap} benchmark, we show:
\vspace{-2mm}
\begin{itemize}[leftmargin=*,topsep=1pt, noitemsep]
\item Pretrained video diffusion models can be directly used as visual trackers.
\item The object permanence capabilities of generative models enable tracking through occlusion.
\item Points can be reliably propagated through video using a novel diffusion prompting strategy.
\item Tracking performance improves through iterative refinement using inpainting.

\item We significantly outperform previous zero-shot tracking methods, such as those that use features from pretrained image diffusion models.
\item Trajectories produced by pretrained video generators can be distilled into a fast tracker with comparable performance.
\end{itemize}
\vspace{-2mm}
We see this work as being a step toward understanding the capabilities of large, pretrained video diffusion models, and new ways to extract these capabilities from them.

\mysection{Related Work}

\vspace{-2mm}
\looseness=-1 \xhdr{Self-supervised Motion Estimation.}
Deep learning has significantly advanced motion estimation. Early dense optical flow methods~\citep{dosovitskiy2015flownet, sun2018pwc, teed2020raft} showed strong performance but often struggle with long-range tracking and occlusions. Inspired by~\citet{sand2008particle}, recent methods instead track individual points over time~\citep{harley2022particle, doersch2022tap}, with newer architectures~\citep{doersch2023tapir, karaev2024cotracker, karaev2024cotracker3, neoral2024mft, zheng2023pointodyssey, doersch2024bootstap, zholus2025tapnext} improving long-term accuracy. However, these models often rely on synthetic data, limiting their real-world generalization. To bridge this gap, self-supervised optical flow methods~\citep{jonschkowski2020matters, liu2019selflow, huang2023self} have been proposed, but they inherit many limitations of supervised approaches. Other work focuses directly on long-range tracking: \citet{Vondrick_2018_ECCV} train a model to propagate color in grayscale videos, implicitly learning motion. Cycle consistency has also been used~\citep{jabri2020space, wang2019learning}, including for point tracking~\citep{shrivastava2024gmrw}.
Models trained for semantic understanding, such as DINOv2~\citep{oquab2023dinov2}, have also been adapted for semantic and temporal correspondence. DIFT~\citep{tang2023emergent}, based on image diffusion models, extracts features suitable for matching, while SD-DINO~\citep{zhang2023tale} combines Stable Diffusion and DINO features to solve a range of semantic and geometric tasks. A recent concurrent work~\citep{nam2025emergent}  extracts features from a pretrained video model for tracking, using a one-to-one frame-to-latent mapping to avoid temporal compression, but involves a complex, architecture-dependent analysis to identify which layers provide the best features and does not handle occlusion. Instead of performing feature extraction, our method prompts a video diffusion model, and thus it is architecture-agnostic. It also handles occlusion by exploiting the ability of modern video diffusion models to successfully convey object permanence, which is not possible with existing methods that work by matching individual pairs of video frames.

\vspace{-2mm}
\xhdr{Counterfactual Modeling.} 
Prior work has explored counterfactual reasoning for visual understanding. Visual Jenga~\citep{bhattad2025visualjengadiscoveringobject} progressively removes objects from a single image until only the background remains, revealing geometric relationships among scene elements. Recent research on counterfactual world modeling~\citep{bear2023unifying,venkatesh2023understanding} tackles keypoint prediction and optical flow by training a masked autoencoder for future-frame prediction, then perturbing inputs to estimate motion. In contrast, we exploit properties of diffusion, such as the ability to subtly manipulate videos, to obtain our predictions from an off-the-shelf model; we base our approach on generative video models rather than masked future frame prediction; and we address the long-range point tracking problem rather than optical flow. Recently, \citet{stojanov2025self} extended the counterfactual world modeling to  point tracking by learning RGB perturbations that can be propagated through a frozen next-frame predictor, optimizing them with a jointly trained sparse optical-flow module. By contrast, our approach relies entirely on prompting a frozen video diffusion model and requires no additional training.

\vspace{-2mm}
\xhdr{Pretrained Models.}
Large pretrained models have become foundational in computer vision, replacing task-specific architectures across classification, detection, and segmentation~\citep{donahue2014decaf, chen2020simple, he2020momentum, zhang2016colorful, oquab2023dinov2, radford2021learning, zhai2023sigmoid, kirillov2023segment, yang2024depth, liu2024grounding, tong2024cambrian, li2023mask}. Diffusion models for image generation~\citep{podell2023sdxl, rombach2022high, dhariwal2021diffusion, nichol2021glide} introduced generative features that capture semantic correspondences~\citep{tang2023emergent, luo2023diffusion, zhang2023tale}, but lack temporal reasoning needed for motion-centric tasks.
Video diffusion models~\citep{Blattmann2023StableVD, Blattmann2023AlignYL, Yu2023VideoPD, wang2025wan, Yang2024CogVideoXTD, Polyak2024MovieGA, Chefer2025VideoJAMJA} address temporal consistency, though many still prioritize appearance over motion.  Recently,~\citet{Chefer2025VideoJAMJA} address this by incorporating optical flow during training. We work in the opposite direction, using generative models to aid motion estimation.

\vspace{-2mm}
\xhdr{Visual Prompting.}
Prompting strategies have achieved notable success in natural language processing~\citep{wei2022chain, kojima2022large}, motivating analogous techniques in computer vision. One prominent direction frames downstream vision tasks as inpainting problems, using pretrained models to complete images conditioned on visual cues~\citep{bar2022visual, wang2023images, bai2024sequential}. Another line of work focuses on optimizing prompt representations, showing that both textual and visual prompts can be refined via gradient-based methods to better adapt vision models~\citep{zhou2022learning, bahng2022exploring}. Recent studies also demonstrate that simple visual prompts, such as colored shapes, can elicit useful behaviors from vision-language models~\citep{shtedritski2023does, yao2024cpt}.  We introduce a simple yet effective visual prompt: placing a colored dot at the pixel to be tracked. To our knowledge, this is the first use of image prompting for point tracking in video diffusion models.

\vspace{-2mm}
\xhdr{Controllable Generation.}
Controllable generation is a key goal in generative modeling~\citep{hao2018controllable, zhuang2021enjoy, liu2021deflocnet, jo2019sc, chen2024exploring, zhang2023adding, ruiz2023dreambooth, chen2023control}. SDEdit~\citep{meng2021sdedit} introduced a training-free method for guided synthesis using noise perturbation and iterative denoising. More recent work enables fine-grained spatial control in diffusion models~\citep{chen2024exploring, lugmayr2022repaint, si2024freeu, wu2024freeinit, chefer2023attend}. RePaint~\citep{lugmayr2022repaint}, for example, inpaints masked regions without affecting the rest of the image. Methods like ControlNet~\citep{zhang2023adding} and DreamBooth~\citep{ruiz2023dreambooth} enable control via fine-tuning.
These ideas have been extended to video~\citep{zhang2023controlvideo,feng2024ccedit}, providing structured editing through architectural design and hierarchical sampling.
\mysection{Method}
Our goal is to repurpose a pretrained generative video model to track points in a video. To do this, we exploit several key properties of diffusion models. We review diffusion models, then describe how they can be adapted for point tracking.

\begin{figure*}[t]
    \centering
    \includegraphics[width=\textwidth]{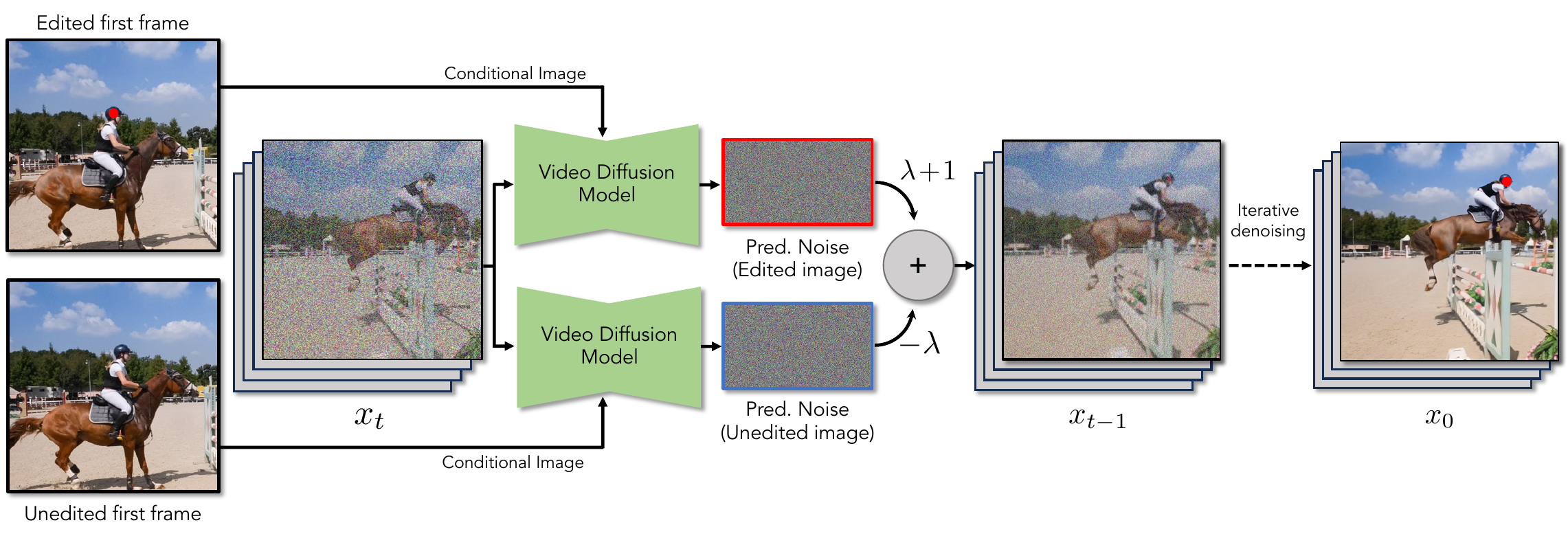}\vspace{-2mm}
\caption{{\bf Enhancing the Counterfactual Signal.}
    We use negative prompting to ensure that the generated video contains the marker. In each denoising step (Eq.~\ref{eq:guidance_ddpm}), we condition the denoising on two images: (1) \textit{Edited First Frame}: the first frame of the video with a marking added, and (2) \textit{Unedited First Frame}: the original first frame of the video. We then subtract the weighted noise vector of the latter from the former.}
    \label{fig:noise_trick}
    \vspace{-3mm}
\end{figure*}

\mysubsection{Preliminaries: Video Diffusion Models}

Latent video diffusion models~\citep{sohl2015deep, ho2020denoising,rombach2022high, Blattmann2023StableVD, wang2025wan} generate a sequence of \(F\) RGB frames, ${\mathbf V} \in \mathbb{R}^{F \times H \times W \times 3}$. These models operate on a compact latent representation ${\mathbf x} \in \mathbb{R}^{F' \times H' \times W' \times C}$, where $C$ is the channel dimension, which can be converted into a video via a decoder.

\vspace{-2mm}
\xhdr{Forward (Noising) Process.\footnote{Our method is agnostic to the specific diffusion model and therefore follows the widely used standard notation of denoising diffusion models~\citep{ho2020denoising} with classifier-free guidance~\citep{ho2022classifier}.}}
Given a clean video latent $\mathbf{x}_0$, we define the noising process using a variance schedule $\beta_t$ over timesteps $t \in \{1,\dots,T\}$. The corrupted latent is constructed via:
\begin{equation}
\mathbf{x}_t = \sqrt{\alpha_t} \mathbf{x}_{t-1} + \sqrt{1-\alpha_t} \boldsymbol{\epsilon}, \quad \boldsymbol{\epsilon} \sim \mathcal{N}(\mathbf{0}, \mathbf{I}),
\label{eq:forward_ddpm}
\end{equation}
where $\alpha_t = 1-\beta_t$ and $\bar{\alpha}_t = \prod_{s=1}^t \alpha_s$. 

\vspace{-2mm}
\paragraph{Reverse (Denoising) Process.}  
At each timestep $t$, the video diffusion model, $\boldsymbol{\epsilon}_\theta(\mathbf{x}_t, t, c)$, predicts the noise component. These models may be conditioned on additional data $c$, such as a text prompt or the desired first frame of the video. We denoise the corrupted latent~\citep{sohl2015deep,ho2020denoising}:
\begin{equation}
\mathbf{x}_{t-1} = \frac{1}{\sqrt{\alpha_t}} \left(\mathbf{x}_t - \frac{\beta_t}{\sqrt{1-\bar{\alpha}_t}} \boldsymbol{\epsilon}_\theta(\mathbf{x}_t, t, c)\right) + \sigma_t \mathbf{z}
\label{eq:reverse_ddpm}
\end{equation}
where $\sigma_t^2$ is the variance, and $\mathbf{z} \sim \mathcal{N}(0, I)$. 

\vspace{-2mm}
\paragraph{Video Manipulation.} Trained diffusion models can also be used to manipulate existing videos, without additional training. We discuss two such applications: regeneration and inpainting.

Rather than generating a latent vector from scratch, one can regenerate an existing, clean video with modifications using SDEdit~\citep{meng2021sdedit}. We add an intermediate level of noise, $1 < t < T$:
\begin{equation}
\mathbf{x}_t = \sqrt{\bar{\alpha}_t} \mathbf{x}_0 + \sqrt{1-\bar{\alpha}_t} \boldsymbol{\epsilon}, 
\label{eq:forward_direct}
\end{equation}
and then run the reverse diffusion process to denoise it. This results in a video that resembles the coarse structure of the original, but with different fine-grained details (e.g., restyling a real video into a cartoon using a text prompt).

We can also use pretrained video diffusion models for inpainting~\citep{lugmayr2022repaint}.
Given a binary spatiotemporal mask \( {\mathbf m} \in \mathbb{B}^{F \times H \times W} \) indicating which patches of the input video can (and cannot) be changed, we run the reverse diffusion process and constrain updates to the masked region. At each denoising step, we constrain the updates such that they occur only in the masked region. In each step of the reverse diffusion process, we compute~\citep{lugmayr2022repaint}:
\begin{equation}
\begin{aligned}
\tilde{\mathbf{x}}_{t-1} &= \frac{1}{\sqrt{\alpha_{t}}} \left(\mathbf{x}_{t} - \frac{\beta_{t}}{\sqrt{1 - \bar{\alpha}_{t}}} \boldsymbol{\epsilon}_\theta\right) + \sigma_{t} \mathbf{z}, \quad \mathbf{z} \sim \mathcal{N}(\mathbf{0}, \mathbf{I}), \\
\mathbf{x}^{\text{original}}_{t-1} &= \sqrt{\bar{\alpha}_{t-1}} \mathbf{x}_0 + \sqrt{1 - \bar{\alpha}_{t-1}} \boldsymbol{\epsilon}, \quad \boldsymbol{\epsilon} \sim \mathcal{N}(\mathbf{0}, \mathbf{I}), \\
\mathbf{x}_{t-1} &= {\mathbf m} \odot \tilde{\mathbf{x}}_{t-1} + (1 - {\mathbf m}) \odot \mathbf{x}^{\text{original}}_{t-1},
\label{eq:repaint}
\end{aligned}
\end{equation}
where $\boldsymbol{\epsilon}_\theta$ is the estimated noise for the  iteration $t$, and as before ${\mathbf x}_0$ is the latent for the  input video.

\mysubsection{Point Prompting for Counterfactual Tracking} \label{sec:background}
We now describe an approach to counterfactual modeling that enables a video diffusion model to perform ``zero shot'' tracking.

\vspace{-2mm}
\paragraph{Marking a Point's Trajectory.} Given an input video and the pixel location of a query point, our goal is to predict the positions of the point in the subsequent frames. As shown in Fig.~\ref{fig:teaser}, we prompt an off-the-shelf video diffusion model to draw a distinctive marker in each frame at the point's position. We then localize the point position using simple low-level image processing. 

We insert a distinctive marking on the query point's position in the initial frame. For this, we simply use a circular dot, which can plausibly be interpreted as being part of the object's surface. For simplicity, we color this dot pure red in all of our experiments. We then apply SDEdit (Sec.~\ref{sec:background}) using an intermediate timestep $1 < t < T$ to the video to manipulate the video, while conditioning on the edited initial frame. This propagates the marker to the subsequent frames of the video.

\begin{figure*}[t]
    \centering
    \includegraphics[width=\textwidth]{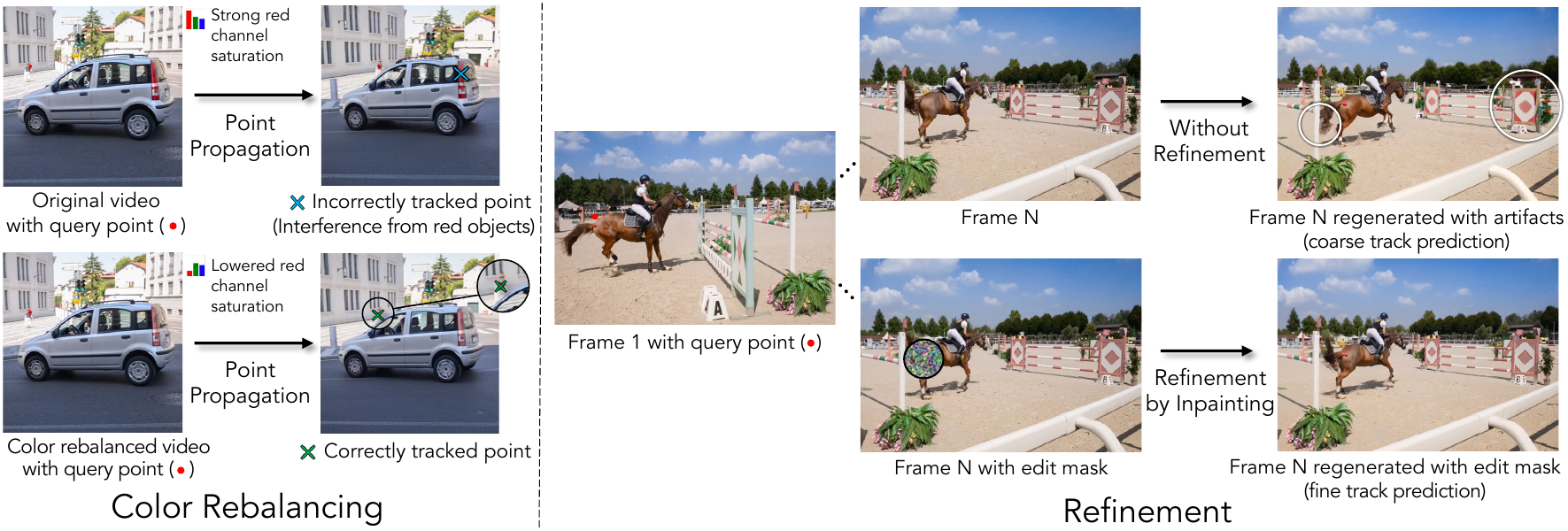}\vspace{-1.5mm}
\caption{{\bf Tracking Enhancements.}
To improve point tracking in video, we introduce two enhancements: (1) \textit{Color Rebalancing:} remove existing red hues to ensure the red marker remains a unique tracking cue; (2) \textit{Refinement:} obtain initial trajectories with a color-based tracker, then refine them using an inpainting mask to correct temporal artifacts such as object shifts (as shown in white circles). This two-step procedure first produces coarse tracks and then refines them via mask-constrained reverse diffusion.
}
    \label{fig:pipeline}
    \vspace{-5mm}
\end{figure*}

\xhdr{Enhancing the Counterfactual Signal.} One of the challenges of applying counterfactual modeling to powerful generative models is that their strong priors lead them to ignore the manipulations that we introduce. For example, when the marker does not naturally fit into a scene, it will often disappear from the generated video within a few frames. We address this problem by using a simple negative prompt that reduces the probability of drawing samples that resemble the original video. We compute the difference between two noise estimates (Fig.~\ref{fig:noise_trick}) that are computed using different types of first-frame conditioning: one where we condition on the original image (i.e., without the marker) and another where we condition on the edited image (i.e., with the marker):
\begin{equation}
   \tilde{\boldsymbol{\epsilon}}_\theta\left(\mathbf{x}_t, \mathbf{c}_{I}\right) = 
    (\lambda + 1) \cdot \boldsymbol{\epsilon}_\theta(\mathbf{x}_t, \phi({\mathbf c_{I}})) -
    \lambda \cdot \boldsymbol{\epsilon}_\theta\left(\mathbf{x}_t, {\mathbf c_{I}}\right),
    \label{eq:guidance_ddpm}
\end{equation} where 
$\tilde{\boldsymbol{\epsilon}}$ is the noise estimate after enhancement, $\mathbf{c}_{I}$ is the initial-frame conditioning, $\phi(\mathbf{c}_{I})$ is the initial frame after applying the counterfactual manipulation (i.e., adding the marker), and $\lambda > 0$ is a weight.  Due to the well-known connection between denoising and score functions, the modified denoiser $\tilde{\boldsymbol{\epsilon}}$ corresponds to the following score function~\citep{ho2022classifier,karras2024guiding}:
\begin{equation}
     \nabla_{{\mathbf x}_t} \log(p_\lambda(\mathbf{x}_t)) 
     =
    \nabla_{\mathbf{x}_t} \log \left(p(\mathbf{x}_t \mid \phi(\mathbf{c}_I)) \left[\frac{p(\phi(\mathbf{c}_I) \mid \mathbf{x}_t)}{p(\mathbf{c}_I \mid \mathbf{x}_t)}\right]^\lambda\right),
\end{equation}
where $p(\mathbf{x}_t)$ is the probability under the model for the noisy input at time $t$, and where we have used the well-known fact that $\boldsymbol{\epsilon}(\mathbf{x}_t) \propto -\nabla_{{\mathbf x}_t} \log(p(\mathbf{x}_t))$ and Bayes rule, following the standard derivation of classifier-free guidance~\citep{ho2022classifier}. From this perspective, we see that our sampling procedure generates videos conditioned on the manipulated initial frame, while biasing the score direction away from samples from the unedited conditioning.

We note that this strategy is related to (but distinct from) the approach used in previous work on counterfactual world models~\citep{bear2023unifying,stojanov2025self}. They generate two possible future images using a masked autoencoder: one with the marker and one without. They then enhance the signal by directly subtracting the two generated images, which amounts to approximately estimating: $\mathbb{E}_{p\left(\mathbf{x} \mid \phi(\mathbf{c}_I)\right)}\left[\mathbf{x}\right] - \mathbb{E}_{p\left(\mathbf{x} \mid \mathbf{c}_I \right)}\left[\mathbf{x}\right].$ Like our approach, this method enhances their ability to detect the effect of the counterfactual by comparing the generated result to an unedited baseline, but instead of comparing the predicted samples themselves, we include this constraint as guidance in the sampler. In our experiments, we found that objects often subtly change position in different samples of a video diffusion model, leading to this differences between generations to contain significant artifacts, making it challenging to use this approach.

\vspace{-1mm}
\xhdr{Tracking the Marker.}
To extract a track from generated videos containing an inserted  marker at a query point, we implement a very simple tracker that locates the marker in each frame based on color. Given the marker’s initial location $(u_0, v_0)$ in the first frame, we track its motion frame by frame. For each subsequent frame $k$, the tracker searches for red pixels (in HSV colorspace) within a local window of radius $r$ centered at the previous location $(u_{k-1}, v_{k-1})$, selecting the pixel closest to the previous position. Since the marker appears as a small blob, we refine the estimate by averaging the positions of nearby red pixels to obtain a more stable center, which serves as the predicted track point.

If no red pixels are found within the search region, we treat the marker as occluded and propagate the last known position forward. We  expand the search radius $r$ at each step until the marker reappears, after which we reset $r$ to its original value. This adaptive strategy makes the tracker robust to temporary occlusions and large displacements, enabling it to recover from tracking uncertainty.

\vspace{-2mm}
\subsection{Extensions}
\vspace{-1mm}
We can further improve the prediction by coarse-to-fine refinement and by rebalancing the colors in the video to exclude the marker's color (Fig.~\ref{fig:pipeline}).

\vspace{-2mm}
\xhdr{Coarse-to-Fine Refinement.}
Accurate tracking requires that the generated video remain pixel-aligned with the original. However, the generated video may be subtly misaligned with the original video after regeneration, leading to tracking errors. Inspired by coarse-to-fine motion estimation, we improve our tracking predictions after their initial estimates, by exploiting the fact that video diffusion models can be repurposed to perform  inpainting. We restrict the model’s ability to modify the video during generation, allowing it to generate only regions near the potential tracked point, while preserving the rest of the video content.

After obtaining the initial estimate of marker positions (as described above), we construct a spatiotemporal binary mask \( {\mathbf m} \in \mathbb{R}^{F \times H \times W} \), where each frame’s mask is set to 1 within a small radius \( r \) centered on the tracked location, i.e., ${\mathbf m}[u, v]$ is set to 1 if $(u, v) \in B_r(u_k, v_k)$.
We then re-run the video generation, while allowing only the image regions indicated by ${\mathbf m}$ to change, using Eq.~\ref{eq:repaint}.

\mypar{Color Rebalancing.}
Since our tracker relies on detecting a particular color, we rebalance the video's colors such that the marker's color does not appear within it. We do this by reducing the saturation of the marker's color. For example, when tracking a red marker, we reduce the saturation of red regions, effectively suppressing natural red hues while preserving overall image quality (details provided in Appendix~\ref{sec:video-preprocess}). We find that this reduces mistakes during occlusion, since the marker is not present and thus false detections are more common.
\vspace{-1mm}
\mysection{Experiments}
\vspace{-1mm}

We evaluate our prompting strategy's ability to accurately track points through a video, using the TAP-Vid benchmarks~\citep{doersch2022tap}.

\mysubsection{Video Models}
\vspace{-1mm}
We consider recent image-conditioned video diffusion models:

\vspace{-2mm}
\xhdr{Wan2.1}~\citep{wang2025wan} combines a 3D causal VAE with a diffusion transformer (DiT) conditioned on text and an input image and trained using flow-matching~\citep{Lipman2022FlowMF}.  The VAE encodes video into latents $x \in \mathbb{R}^{(1+F/4) \times H/8 \times W/8}$, keeping the first frame at full temporal resolution and downsampling the rest by $4\times$. Outputs are $480 \times 832$.
We test 1.3B- and 14B-parameter variants, reporting results with the 14B model unless noted.

\vspace{-2mm}
\xhdr{Wan2.2}~\citep{wang2025wan} extends Wan2.1 with a Mixture-of-Experts (MoE) architecture.
By distributing denoising across timesteps among specialized experts, it increases model capacity without extra computation and is trained on a much larger dataset.

\vspace{-2mm}
\xhdr{CogVideoX}~\citep{Yang2024CogVideoXTD} is another image-to-video (I2V) diffusion model that also combines a 3D causal VAE with a diffusion transformer.
It generates $768 \times 1360$ videos from a text prompt and reference image.
The VAE compresssion is the same as Wan, while the transformer conditions on the image and T5 text embeddings~\citep{raffel2020exploring}.

For all models, we use 50 denoising steps with noise strength 0.5 and an empty text prompt.
Experiments run on A40 or L40S GPUs (one GPU per video).
Generating a 50-frame video for a single query point takes about 7 min for Wan2.1-1.3B, 30 min for Wan2.1-14B, and 20 min for CogVideoX.
These runtimes are acceptable given our focus on evaluating the tracking capabilities of video diffusion models. We note that method could potentially be distilled into a more efficient model, similar to Opt-CWM~\citep{stojanov2025self}.

\vspace{-2mm}

\subsection{TAP-Vid Benchmark}
We evaluate on two TAP-Vid benchmark splits: DAVIS (30 videos, 34–104 frames) and Kinetics (30 sampled videos, 250 frames, following~\citep{stojanov2025self}) for efficiency. These natural videos match the training distribution of our video diffusion models (rather than computer generated video). Using the first sampling strategy, we pick one query point per video, overlay a red dot at its position in the first frame, and run our model to propagate the point throughout the video. The resulting trajectory is then extracted using our tracker.

\xhdr{Evaluation Metrics.}
We report: (1) \emph{Positional Accuracy} ($\delta_{\text{avg}}^{x}$), fraction of visible points within distance thresholds; (2) \emph{Occlusion Accuracy} (OA), visibility prediction accuracy; and (3) \emph{Average Jaccard} (AJ), average overlap between predicted and ground-truth visible points across thresholds~\citep{doersch2022tap}.

\mysection{Results}
\label{sec:results}

Unless otherwise noted, we use Wan2.1-14B~\citep{wang2025wan} as the video diffusion model for all experiments.

\begin{table*}[h!]
\vspace{-2mm}
\small
\centering
{\resizebox{\linewidth}{!}{

\begin{tabular}{ll ccc ccc}
\toprule
\multirow{2}{*}{\textbf{Method}} &
\multirow{2}{*}{\textbf{Supervision}} &
\multicolumn{3}{c}{\textbf{TAP-Vid DAVIS}} &
\multicolumn{3}{c}{\textbf{TAP-Vid Kinetics}}
\\

\cmidrule(lr){3-5} \cmidrule(lr){6-8}

& & AJ~$\uparrow$ & $<\delta^x_\textrm{avg}$~$\uparrow$ & OA~$\uparrow$ &
AJ~$\uparrow$ & $<\delta^x_\textrm{avg}$~$\uparrow$ & OA~$\uparrow$ \\
\midrule
RAFT~\citep{teed2020raft}      & \multirow{5}{*}{Supervised} & $34.48$ & $53.55$ & $74.90$ & $30.15$       & $46.44$     & $75.44$    \\
TAP-Net~\citep{doersch2022tap}   &  & $32.05$ & $48.42$ & $77.35$ & $34.59$ & $48.42$ & $80.88$   \\

TAPIR~\citep{doersch2023tapir}     &  & $58.47$ & $70.56$ & $87.27$ & $47.46$       & $59.56$     & $85.76$ \\
CoTracker3~\citep{karaev24cotracker3}       &  & $64.45$     & $77.13$     & $90.90$     & $\bf{54.35}$       & $\bf{65.99}$     & $\bf{89.43}$  \\
TAPNext~\citep{zholus2025tapnext}           &  & $\bf{66.56}$     & $\bf{79.48}$     & $\bf{92.21}$     & $52.97$       & $64.46$     & $89.30$  \\

\midrule

GMRW~\citep{shrivastava2024gmrw}      & \multirow{2}{*}{Self-Sup.}        & $36.47$ & $54.59$ & $76.36$ & $25.70$   & $41.63$ & $71.33$ \\
Opt-CWM~\citep{stojanov2025self}           &        & $\bf{47.53}$ & $\bf{64.83}$ & $\bf{80.87}$ & $\bf{44.85}$   & $\bf{57.74}$ & $\bf{84.12}$ \\

\midrule

DINOv2+NN~\citep{oquab2023dinov2} &\multirow{4}{*}{Zero-Shot}  & $15.19$     & $31.19$     & $61.81$     & $12.69$       & $24.22$ & $62.45$  \\
DIFT~\citep{tang2023emergent}      &   & $21.51$ & $39.55$ & $69.71$ & $15.10$       & $25.56$     & $63.17$ \\
SD-DINO~\citep{zhang2023tale}   &   & $29.68$     & $50.45$     & $69.71$     & $16.47$       & $28.37$     & $62.79$ \\
Ours              &   & $\bf{42.21}$ & $\bf{57.29}$ & $\bf{82.90}$ & $\bf{27.36}$       & $\bf{41.51}$     & $\bf{71.39}$  \\

\bottomrule
\end{tabular}
}}
\vspace{-2mm}
\caption{{\bf TAP-Vid Benchmark Results.}
We report results on the TAP-Vid First benchmark. Our zero-shot  method outperforms all other zero-shot baselines and is competitive with self-supervised and supervised trackers. On TAP-Vid DAVIS-First, it matches self-supervised methods in AJ and exceeds them in occlusion accuracy, highlighting strong object permanence from generative modeling.}

\label{tab:main_results}
\end{table*}

\vspace{-2mm}

\xhdr{Quantitative Results.}
Table~\ref{tab:main_results} compares our method against several baselines using Wan2.1. Among zero-shot methods, ours achieves the highest performance. On TAP-Vid DAVIS, we reach an AJ score of 42.21, outperforming all other zero-shot baselines and even surpassing GMRW~\citep{shrivastava2024gmrw}, a strong self-supervised approach. Our occlusion accuracy also exceeds that of both zero-shot and self-supervised methods, approaching supervised performance, highlighting the ability of diffusion models to reason through occlusions.

We include top supervised methods such as CoTracker3~\citep{karaev24cotracker3} and TAPNext~\citep{zholus2025tapnext}, as well as the best-performing self-supervised baseline, Opt-CWM~\citep{stojanov2025self}. While conceptually related, Opt-CWM learns to propagate perturbations through a next-frame predictor supervised by sparse flow. In contrast, our method is entirely zero-shot, using a simple colored dot without training or learned perturbations. As shown in Table~\ref{tab:kubric}, we also report results on TAP-Vid Kubric, where overall performance is comparatively lower, likely due to the dataset’s synthetic nature and the fact that existing video models are primarily trained on real videos.

\vspace{-2mm}
\begin{figure}[h!]
  \centering
  \begin{minipage}[b]{0.58\linewidth}
    \centering
    \resizebox{\linewidth}{!}{
        \setlength{\tabcolsep}{2pt}
\begin{tabular}{l ccc}
\toprule
\multirow{2}{*}{\textbf{Method}} &
\multicolumn{3}{c}{\textbf{TAP-Vid DAVIS}}
\\

\cmidrule(lr){2-4}
& AJ~$\uparrow$ & $<\delta^x_\textrm{avg}$~$\uparrow$ & OA~$\uparrow$
\\

\midrule

CogVideoX1.5-5B~\citep{Yang2024CogVideoXTD} & 24.15 & 34.38 & 70.79 \\
Wan2.1-1.3B~\citep{wang2025wan}         & 44.58 & 58.77 & 85.16 \\
Wan2.1-14B~\citep{wang2025wan}          & 48.60 & 63.47 & 85.75 \\
Wan2.2-14B~\citep{wang2025wan}          & 48.78 & 63.91 & 86.17 \\
\bottomrule
\end{tabular}
    }
    \vspace{-2mm}
    \captionof{table}{{\bf Video Model Ablations. }{Wan2.1-1.3B and 14B~\citep{wang2025wan} outperform CogVideoX~\citep{Yang2024CogVideoXTD}, showing that stronger video models improve tracking performance.}}
    \label{tab:ablations_video_models}
  \end{minipage}\hfill
  \vspace{-2mm}
  \begin{minipage}[b]{0.4\linewidth}
    \centering
    \resizebox{\linewidth}{!}{
      \setlength{\tabcolsep}{2pt}
\begin{tabular}{l ccc}
\toprule
\multirow{2}{*}{\textbf{Image source}} &
\multicolumn{3}{c}{\textbf{TAP-Vid DAVIS}}
\\

\cmidrule(lr){2-4}

& AJ~$\uparrow$ & $<\delta^x_\textrm{avg}$~$\uparrow$ & OA~$\uparrow$ \\

\midrule
DAVIS (256$\times$256) & 42.21 & 57.29 & 82.90 \\
DAVIS (256$\times$256 up.) & 45.48 & 60.16 & 83.49 \\
DAVIS (original res.) & 48.60 & 63.47 & 85.75 \\

\bottomrule
\end{tabular}
    }
    \vspace{-2mm}
    \captionof{table}{{\bf Image Resolution Ablations.} Comparing input resolutions for Wan2.1. Upscaling with~\citep{zhou2024upscale} improves tracking by better aligning with the model’s training distribution.}
    \label{tab:different_res}
  \end{minipage}
\end{figure}

\vspace{2mm}
\begin{figure}[h!]
  \begin{minipage}[b]{.6\linewidth}
    \centering
    \raisebox{26.5mm}{
      \parbox{.95\linewidth}{
        \centering
        \resizebox{\linewidth}{!}{
          
\setlength{\tabcolsep}{6pt}
\begin{tabular}{l ccc}

            \toprule
            \multirow{2}{*}{\textbf{Method}} &
            \multicolumn{3}{c}{\textbf{TAP-Vid DAVIS}}
            \\
            \cmidrule(lr){2-4}
            & AJ~$\uparrow$ & $<\delta^x_\textrm{avg}$~$\uparrow$ & OA~$\uparrow$ \\
            \midrule
            all     & 48.60 & 63.47 & 85.75 \\
            w/o refinement & 42.70 & 59.26 & 85.14 \\
            w/o counterfactual enhancement & 22.03 & 38.53 & 61.19 \\
            w/o color rebalancing & 34.86 & 52.12 & 82.18 \\
            tracker only & 11.26 & 21.07 & 77.74 \\
            \bottomrule
\end{tabular}

        }
        \vspace{-2mm}
        \captionof{table}{{\small \bf Tracking Pipeline Ablations.} Quantitative results on TAP-Vid DAVIS-First showing the impact of each stage in our pipeline (Fig.~\ref{fig:pipeline}). The last row uses original pixel color instead of the red dot for tracking.}
        \label{tab:ablations_video}
      }
    }
    \vspace{-4mm}
  \end{minipage}\hfill
  \begin{minipage}[b]{.35\linewidth}
    \centering
    \includegraphics[width=\linewidth]{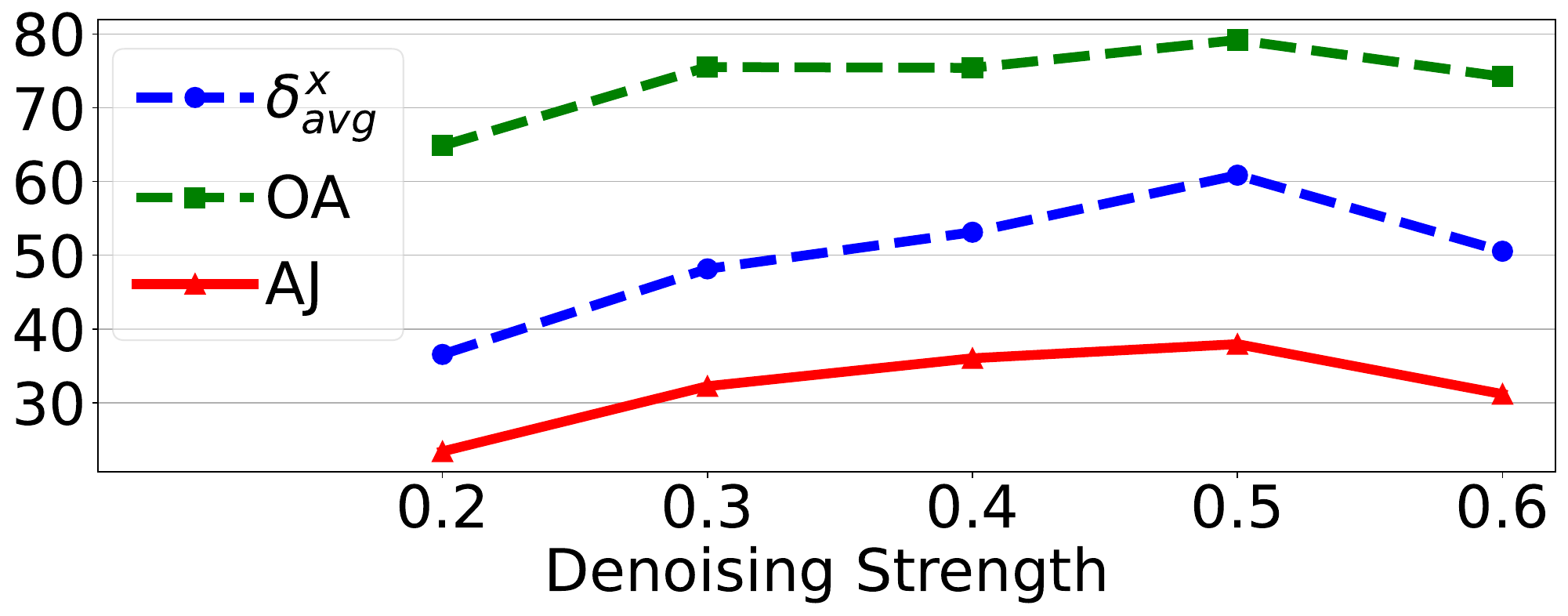}
    \vspace{-3mm}
    \label{fig:denoising_performance}

    \vspace{-1mm}
    \includegraphics[width=\linewidth]{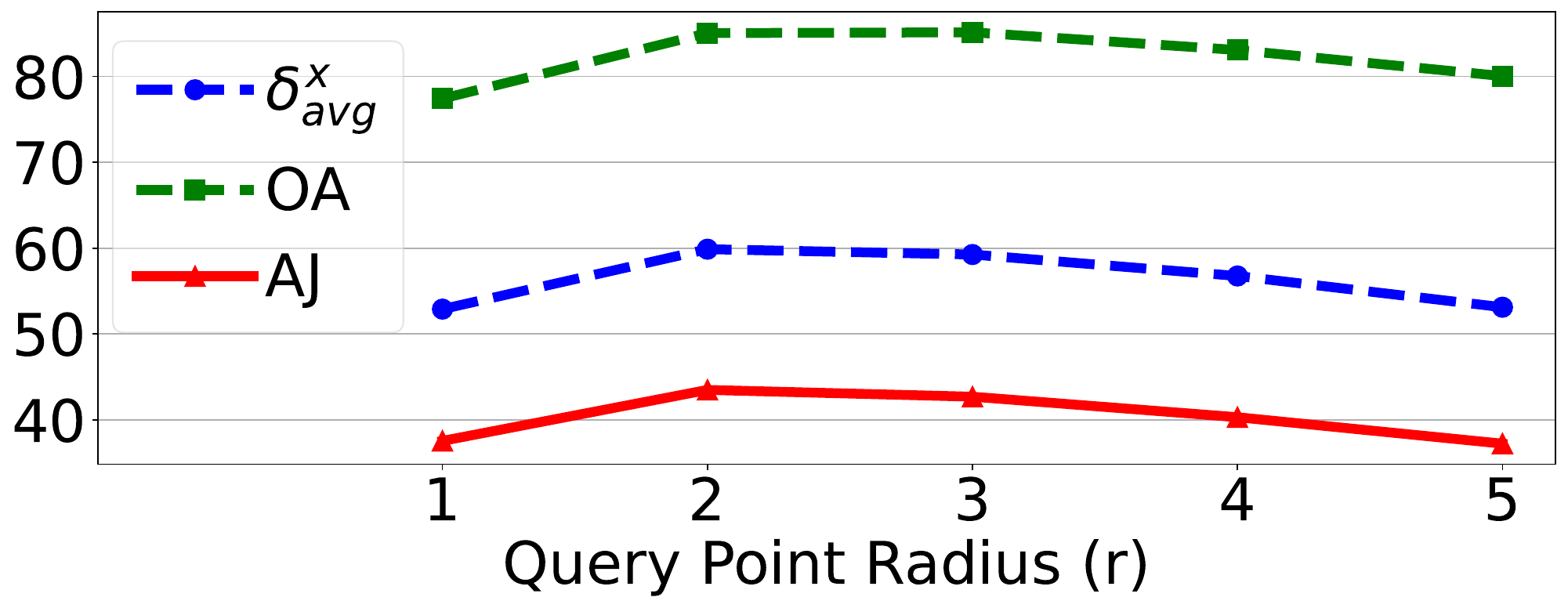}
    \vspace{-6mm}
    \captionof{figure}{{\small \bf Effect of denoising strength and radius on tracking performance.} }
    \label{fig:qp_r_performance}
  \end{minipage}
  \label{fig:radius_performance}
  \vspace{-2mm}
\end{figure}

\vspace{-2mm}

\looseness=-1 
\xhdr{Different Video Models.}
Table~\ref{tab:ablations_video_models} shows results using Wan2.1 (1.3B and 14B variants), Wan2.2, and CogVideoX~\citep{Yang2024CogVideoXTD}. Our method successfully tracks points for all four models, demonstrating compatibility across different video generation backbones. Wan2.1 and Wan2.2 obtain the strongest results, with the 14B variant outperforming the 1.3B model.
We attribute this gain to their higher video generation quality, suggesting that improved generative fidelity directly may improved tracking ability.

\begin{figure*}[t]
    \centering
    \includegraphics[width=\textwidth]{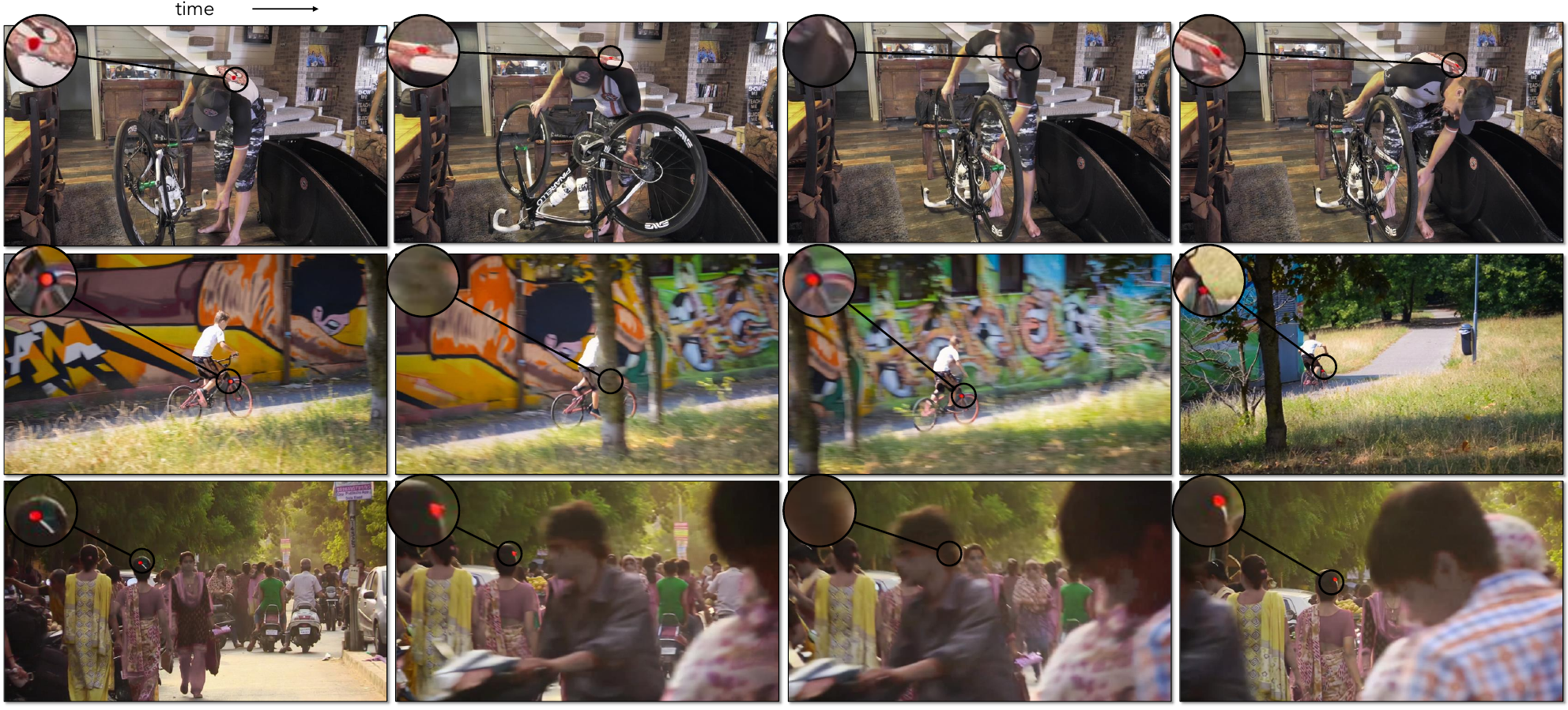}\vspace{-1.5mm}
    \caption{{\bf Point Propagation.} Frames generated from the video diffusion model show consistent red dot tracking. The model recovers the point after long occlusions, showing temporal understanding and object permanence.}
    \label{fig:qual_fig1}
    \vspace{-4mm}
\end{figure*}

\vspace{-2mm}
\xhdr{Generation Resolution.}
The TAP-Vid benchmark provides videos at a resolution of 256×256, which we resize to 480×832 to match the input resolution of Wan2.1.
To assess the impact of resolution, we first upsample inputs using Upsample-A-Video~\citep{zhou2024upscale}, which improves tracking (Table~\ref{tab:different_res}). We then run Wan2.1 on the original high-res DAVIS frames~\citep{perazzi2016benchmark}, achieving an AJ score of 48.6, surpassing Opt-CWM. These results show that higher-resolution inputs significantly enhance tracking by improving video generation quality.

\vspace{-2mm}
\xhdr{Point Propagation Ablations.}
Table~\ref{tab:ablations_video} shows ablations of key components. The first row shows our full model with all components enabled. Removing the inpainting-based refinement step reduces positional accuracy due to spatial shifts during denoising which negatively affects tracking precision. Removing counterfactual enhancement guidance causes failure in point propagation where tracking is lost after 5–6 frames, highlighting its critical role in maintaining point consistency across frames. Disabling color rebalancing also degrades performance. Since the tracker relies on detecting red pixels, failure to suppress red tones in the background introduces false positives, especially when the query point is occluded, making tracking less reliable.

We also evaluate a tracker-only baseline that tracks the query point’s color from the initial frame without any point propagation. This performs significantly worse, highlighting that the primary performance gains in our method arise from accurate point propagation through video generation, rather than from the tracker itself, which is intentionally kept simple. Additionally, we ablate key hyperparameters in Fig.~\ref{fig:qp_r_performance}. We observe that a noise strength of 0.5 and a query point radius of 2 pixels yield the best results. The effect of varying the marker color is further analyzed in Table~\ref{tab:marker-color-ablation}.

\vspace{-1mm}
\begin{wraptable}{r}{0.4\linewidth}
\vspace{-3mm}
\setlength{\tabcolsep}{2pt}
\centering
\footnotesize
\begin{tabular}{l ccc}
\toprule
\multirow{2}{*}{\textbf{Method}} &
\multicolumn{3}{c}{\textbf{TAP-Vid DAVIS}} \\
\cmidrule(lr){2-4}
& AJ~$\uparrow$ & $<\delta^x_\textrm{avg}$~$\uparrow$ & OA~$\uparrow$ \\
\midrule
Wan2.1-1.3B (teacher) & 38.78 & 54.75 & 85.00 \\
Cotracker (distilled) & 37.17 & 53.12 & 84.24 \\
\bottomrule
\label{tab:distillation}
\end{tabular}
\vspace{-5mm}
\caption{\textbf{Distilling the generator.} We distill our method to CoTracker which achieves performance close to the teacher Wan2.1 and runs orders of magnitude faster.}
\label{tab:distillation}
\vspace{-4mm}
\end{wraptable}
\xhdr{Distilling the Generator into a Tracker.} 
While our approach shows strong zero-shot tracking performance, it requires a separate video generation for each query point, which limits efficiency. To obtain a fast tracker, inspired by this~\citep{stojanov2025self}, we distill our generation-based method into an efficient tracking architecture. Specifically, we collect pseudo-label trajectories by running our marker propagation method on 1,000 unlabeled videos from the TAP-Vid Kinetics dataset. These extracted trajectories serve as supervision to train CoTracker~\citep{karaev23cotracker} from scratch. 

Table~\ref{tab:distillation} compares the teacher model used to generate trajectories with the distilled CoTracker model. Despite being trained solely on pseudo-labels produced by our zero-shot method, the distilled tracker achieves performance very close to the teacher while running orders of magnitude faster at inference. This indicates that the temporal reasoning capabilities of large video diffusion models can be transferred into a lightweight tracking network, effectively bridging generative zero-shot tracking and efficient feed-forward inference.

\begin{figure*}[t]
    \centering
    \includegraphics[width=\textwidth]{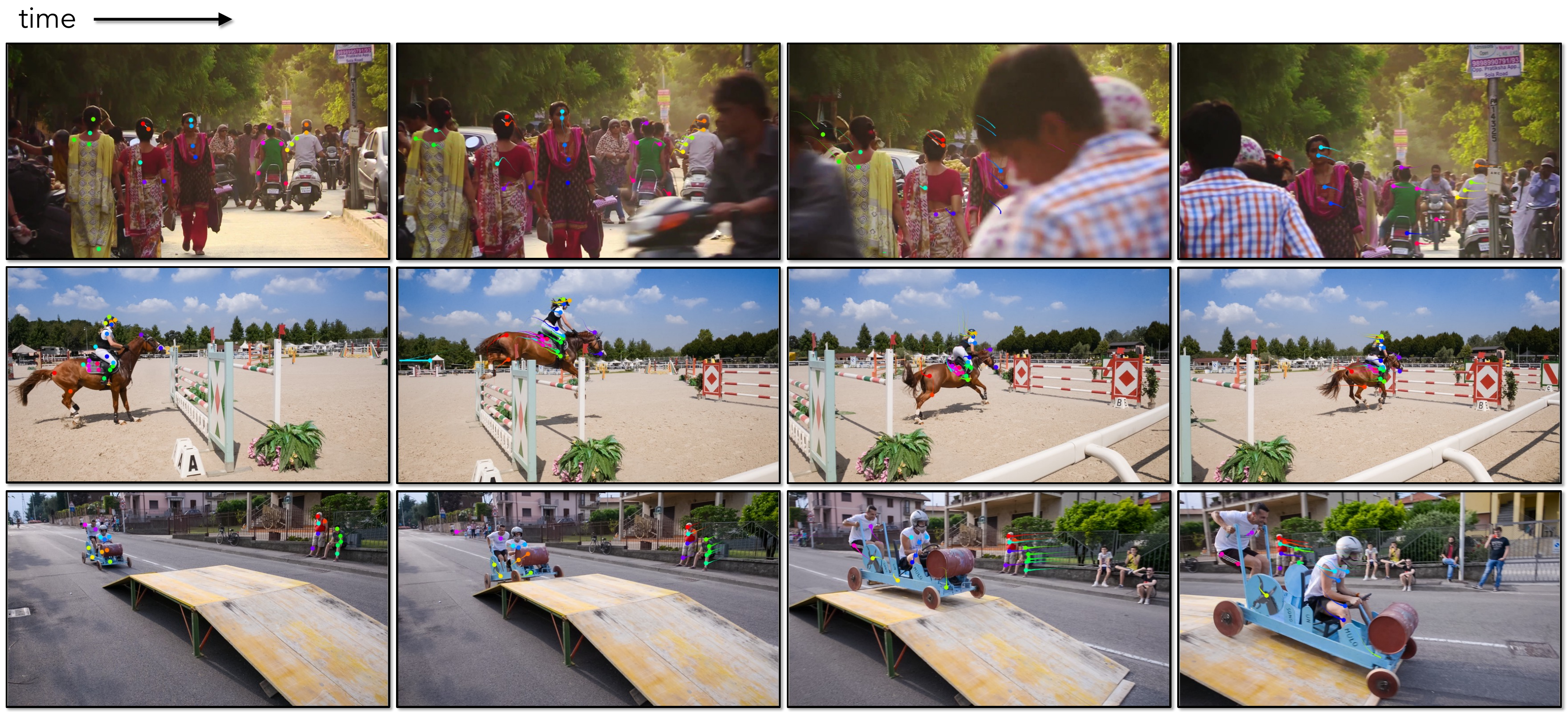}\vspace{-2mm}
    \caption{{\bf \footnotesize Tracking results.} Frames show the query point being tracked (circled dot) and its trajectory over the previous 5 frames. When the query point is occluded, only the trajectory tail is displayed without the dot.}
    \label{fig:tracking_fig}
    \vspace{-5mm}
\end{figure*}

\vspace{-2mm}
\xhdr{Qualitative Results.}
In Fig.~\ref{fig:qual_fig1}, we show video generations from our method, where red dots are successfully propagated across frames, including through occlusions. We extract these points and display the resulting tracks for multiple query points in Fig.~\ref{fig:tracking_fig}. Our method reliably tracks points over long temporal range and maintains accuracy even in the presence of occlusions.

\section{Limitations}
\begin{figure*}[h]
    \centering
    \includegraphics[width=\textwidth]{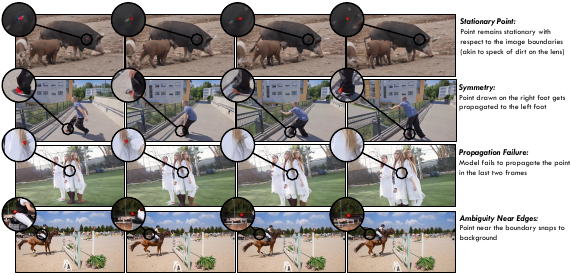}\vspace{-2mm}
    \caption{{\bf Generation Failures.} Typical failure cases in video generation: (1) \textit{Stationary Point:} The red dot remains fixed relative to image boundaries, resembling lens dirt. (2) \textit{Symmetry Confusion:} Symmetrical objects (e.g., left and right body parts) cause point propagation errors, likely due to compressed latent representations. (3) \textit{Propagation Failure:} The red dot vanishes across consecutive frames. (4) \textit{Edge Ambiguity:} The red dot, near boundaries, shifts to the background.}
    \label{fig:generation_failures}
    \vspace{-5mm}
\end{figure*}

Our approach requires generating a video for each tracked point. Since our goal is to show that video generators can perform tracking, rather than to perform tracking as an end in itself, we did not attempt to optimize our approach. However, it can potentially be addressed by distilling our model's predictions into a network that directly performs tracking, by considering more efficient generation methods (e.g., one-step sampling), or by tracking multiple points at once. The video generators also sometimes fail to interpret the red dot as being attached to the object surface, especially for (likely out-of-distribution) computer-generated videos (Fig.~\ref{fig:generation_failures}).

\section{Conclusion}
\label{sec:conclusion}
We have shown that a video diffusion model, when carefully prompted, can mark the location of a point as it moves through a scene over time. We use this idea to create a simple point tracker, which obtains surprisingly effective tracking results, outperforming previous zero-shot approaches. We see our work as opening two new directions. The first is expanding the number of ways that one can adapt large pretrained video diffusion models to new tasks, such as through prompting schemes that go beyond the use of language. Second, our work shows that video generative models are a useful source of pretraining for tracking. We therefore see our work as a step toward unifying video generation and tracking. 
\vspace{-2mm}

\section{Acknowledgements}
\looseness=-1
We would like to thank Adam Harley, Dan Yamins, and Xuanchen Lu for helpful discussions and feedback on the paper. Daniel Geng was at the University of Michigan when he contributed to the project. This work was supported by the Advanced Research Projects Agency for Health (ARPA-H) under award \#1AY2AX000062. This research was funded, in part, by the U.S. Government. The views and conclusions contained in this document are those of the authors and should not be interpreted as representing the official policies, either expressed or implied, of the U.S. Government.

\bibliographystyle{iclr_template/iclr2026_conference}
\bibliography{main}

\begin{thebibliography}{93}
\providecommand{\natexlab}[1]{#1}
\providecommand{\url}[1]{\texttt{#1}}
\expandafter\ifx\csname urlstyle\endcsname\relax
  \providecommand{\doi}[1]{doi: #1}\else
  \providecommand{\doi}{doi: \begingroup \urlstyle{rm}\Url}\fi

\bibitem[Allen et~al.(2025)Allen, Doersch, Zhou, Suhail, Driess, Rocco, Rubanova, Kipf, Sajjadi, Murphy, et~al.]{allen2025direct}
Kelsey Allen, Carl Doersch, Guangyao Zhou, Mohammed Suhail, Danny Driess, Ignacio Rocco, Yulia Rubanova, Thomas Kipf, Mehdi~SM Sajjadi, Kevin Murphy, et~al.
\newblock Direct motion models for assessing generated videos.
\newblock \emph{arXiv preprint arXiv:2505.00209}, 2025.

\bibitem[Ardino et~al.(2021)Ardino, De~Nadai, Lepri, Ricci, and Lathuili{\`e}re]{ardino2021click}
Pierfrancesco Ardino, Marco De~Nadai, Bruno Lepri, Elisa Ricci, and St{\'e}phane Lathuili{\`e}re.
\newblock Click to move: Controlling video generation with sparse motion.
\newblock In \emph{Proceedings of the IEEE/CVF International Conference on Computer Vision}, 2021.

\bibitem[Bahng et~al.(2022)Bahng, Jahanian, Sankaranarayanan, and Isola]{bahng2022exploring}
Hyojin Bahng, Ali Jahanian, Swami Sankaranarayanan, and Phillip Isola.
\newblock Exploring visual prompts for adapting large-scale models.
\newblock \emph{arXiv preprint arXiv:2203.17274}, 2022.

\bibitem[Bai et~al.(2024)Bai, Geng, Mangalam, Bar, Yuille, Darrell, Malik, and Efros]{bai2024sequential}
Yutong Bai, Xinyang Geng, Karttikeya Mangalam, Amir Bar, Alan~L Yuille, Trevor Darrell, Jitendra Malik, and Alexei~A Efros.
\newblock Sequential modeling enables scalable learning for large vision models.
\newblock In \emph{Proceedings of the IEEE/CVF Conference on Computer Vision and Pattern Recognition}, pp.\  22861--22872, 2024.

\bibitem[Bar et~al.(2022)Bar, Gandelsman, Darrell, Globerson, and Efros]{bar2022visual}
Amir Bar, Yossi Gandelsman, Trevor Darrell, Amir Globerson, and Alexei Efros.
\newblock Visual prompting via image inpainting.
\newblock \emph{Advances in Neural Information Processing Systems}, 35:\penalty0 25005--25017, 2022.

\bibitem[Bear et~al.(2023)Bear, Feigelis, Chen, Lee, Venkatesh, Kotar, Durango, and Yamins]{bear2023unifying}
Daniel~M Bear, Kevin Feigelis, Honglin Chen, Wanhee Lee, Rahul Venkatesh, Klemen Kotar, Alex Durango, and Daniel~LK Yamins.
\newblock Unifying (machine) vision via counterfactual world modeling.
\newblock \emph{arXiv preprint arXiv:2306.01828}, 2023.

\bibitem[Bhattad et~al.(2025)Bhattad, Preechakul, and Efros]{bhattad2025visualjengadiscoveringobject}
Anand Bhattad, Konpat Preechakul, and Alexei~A. Efros.
\newblock Visual jenga: Discovering object dependencies via counterfactual inpainting, 2025.
\newblock URL \url{https://arxiv.org/abs/2503.21770}.

\bibitem[Blattmann et~al.(2023{\natexlab{a}})Blattmann, Dockhorn, Kulal, Mendelevitch, Kilian, and Lorenz]{Blattmann2023StableVD}
A.~Blattmann, Tim Dockhorn, Sumith Kulal, Daniel Mendelevitch, Maciej Kilian, and Dominik Lorenz.
\newblock Stable video diffusion: Scaling latent video diffusion models to large datasets.
\newblock \emph{ArXiv}, 2023{\natexlab{a}}.

\bibitem[Blattmann et~al.(2023{\natexlab{b}})Blattmann, Rombach, Ling, Dockhorn, Kim, Fidler, and Kreis]{Blattmann2023AlignYL}
A.~Blattmann, Robin Rombach, Huan Ling, Tim Dockhorn, Seung~Wook Kim, Sanja Fidler, and Karsten Kreis.
\newblock Align your latents: High-resolution video synthesis with latent diffusion models.
\newblock \emph{2023 IEEE/CVF Conference on Computer Vision and Pattern Recognition (CVPR)}, 2023{\natexlab{b}}.

\bibitem[Burgert et~al.(2025)Burgert, Xu, Xian, Pilarski, Clausen, He, Ma, Deng, Li, Mousavi, et~al.]{burgert2025go}
Ryan Burgert, Yuancheng Xu, Wenqi Xian, Oliver Pilarski, Pascal Clausen, Mingming He, Li~Ma, Yitong Deng, Lingxiao Li, Mohsen Mousavi, et~al.
\newblock Go-with-the-flow: Motion-controllable video diffusion models using real-time warped noise.
\newblock \emph{arXiv preprint arXiv:2501.08331}, 2025.

\bibitem[Ceylan et~al.(2023)Ceylan, Huang, and Mitra]{ceylan2023pix2video}
Duygu Ceylan, Chun-Hao Huang, and Niloy~J. Mitra.
\newblock Pix2video: Video editing using image diffusion.
\newblock In \emph{ICCV}, 2023.

\bibitem[Chefer et~al.(2023)Chefer, Alaluf, Vinker, Wolf, and Cohen-Or]{chefer2023attend}
Hila Chefer, Yuval Alaluf, Yael Vinker, Lior Wolf, and Daniel Cohen-Or.
\newblock Attend-and-excite: Attention-based semantic guidance for text-to-image diffusion models.
\newblock \emph{ACM transactions on Graphics (TOG)}, 42\penalty0 (4):\penalty0 1--10, 2023.

\bibitem[Chefer et~al.(2025)Chefer, Singer, Zohar, Kirstain, Polyak, Taigman, Wolf, and Sheynin]{Chefer2025VideoJAMJA}
Hila Chefer, Uriel Singer, Amit Zohar, Yuval Kirstain, Adam Polyak, Yaniv Taigman, Lior Wolf, and Shelly Sheynin.
\newblock Videojam: Joint appearance-motion representations for enhanced motion generation in video models.
\newblock \emph{ArXiv}, 2025.

\bibitem[Chen et~al.(2024)Chen, Zhang, Guo, Lu, Wang, and Qu]{chen2024exploring}
Siyi Chen, Huijie Zhang, Minzhe Guo, Yifu Lu, Peng Wang, and Qing Qu.
\newblock Exploring low-dimensional subspaces in diffusion models for controllable image editing.
\newblock \emph{arXiv preprint arXiv:2409.02374}, 2024.

\bibitem[Chen et~al.(2020)Chen, Kornblith, Norouzi, and Hinton]{chen2020simple}
Ting Chen, Simon Kornblith, Mohammad Norouzi, and Geoffrey Hinton.
\newblock A simple framework for contrastive learning of visual representations.
\newblock In \emph{International conference on machine learning}, pp.\  1597--1607. PmLR, 2020.

\bibitem[Chen et~al.(2023)Chen, Ji, Wu, Wu, Xie, Li, Xia, Xiao, and Lin]{chen2023control}
Weifeng Chen, Yatai Ji, Jie Wu, Hefeng Wu, Pan Xie, Jiashi Li, Xin Xia, Xuefeng Xiao, and Liang Lin.
\newblock Control-a-video: Controllable text-to-video diffusion models with motion prior and reward feedback learning.
\newblock \emph{arXiv preprint arXiv:2305.13840}, 2023.

\bibitem[Dhariwal \& Nichol(2021)Dhariwal and Nichol]{dhariwal2021diffusion}
Prafulla Dhariwal and Alexander Nichol.
\newblock Diffusion models beat gans on image synthesis.
\newblock \emph{Advances in neural information processing systems}, 34:\penalty0 8780--8794, 2021.

\bibitem[Doersch et~al.(2022)Doersch, Gupta, Markeeva, Recasens, Smaira, Aytar, Carreira, Zisserman, and Yang]{doersch2022tap}
Carl Doersch, Ankush Gupta, Larisa Markeeva, Adria Recasens, Lucas Smaira, Yusuf Aytar, Joao Carreira, Andrew Zisserman, and Yi~Yang.
\newblock {TAP}-vid: A benchmark for tracking any point in a video.
\newblock \emph{Advances in Neural Information Processing Systems}, 35:\penalty0 13610--13626, 2022.

\bibitem[Doersch et~al.(2023)Doersch, Yang, Vecerik, Gokay, Gupta, Aytar, Carreira, and Zisserman]{doersch2023tapir}
Carl Doersch, Yi~Yang, Mel Vecerik, Dilara Gokay, Ankush Gupta, Yusuf Aytar, Joao Carreira, and Andrew Zisserman.
\newblock Tapir: Tracking any point with per-frame initialization and temporal refinement.
\newblock In \emph{Proceedings of the IEEE/CVF International Conference on Computer Vision}, pp.\  10061--10072, 2023.

\bibitem[Doersch et~al.(2024)Doersch, Luc, Yang, Gokay, Koppula, Gupta, Heyward, Rocco, Goroshin, Carreira, et~al.]{doersch2024bootstap}
Carl Doersch, Pauline Luc, Yi~Yang, Dilara Gokay, Skanda Koppula, Ankush Gupta, Joseph Heyward, Ignacio Rocco, Ross Goroshin, Jo{\~a}o Carreira, et~al.
\newblock Bootstap: Bootstrapped training for tracking-any-point.
\newblock In \emph{Proceedings of the Asian Conference on Computer Vision}, pp.\  3257--3274, 2024.

\bibitem[Donahue et~al.(2014)Donahue, Jia, Vinyals, Hoffman, Zhang, Tzeng, and Darrell]{donahue2014decaf}
Jeff Donahue, Yangqing Jia, Oriol Vinyals, Judy Hoffman, Ning Zhang, Eric Tzeng, and Trevor Darrell.
\newblock Decaf: A deep convolutional activation feature for generic visual recognition.
\newblock In \emph{International conference on machine learning}, pp.\  647--655. PMLR, 2014.

\bibitem[Dosovitskiy et~al.(2015)Dosovitskiy, Fischer, Ilg, Hausser, Hazirbas, Golkov, Van Der~Smagt, Cremers, and Brox]{dosovitskiy2015flownet}
Alexey Dosovitskiy, Philipp Fischer, Eddy Ilg, Philip Hausser, Caner Hazirbas, Vladimir Golkov, Patrick Van Der~Smagt, Daniel Cremers, and Thomas Brox.
\newblock Flownet: Learning optical flow with convolutional networks.
\newblock In \emph{Proceedings of the IEEE international conference on computer vision}, pp.\  2758--2766, 2015.

\bibitem[Feng et~al.(2024)Feng, Weng, Wang, Yuan, Bao, Luo, Chen, and Guo]{feng2024ccedit}
Ruoyu Feng, Wenming Weng, Yanhui Wang, Yuhui Yuan, Jianmin Bao, Chong Luo, Zhibo Chen, and Baining Guo.
\newblock Ccedit: Creative and controllable video editing via diffusion models.
\newblock In \emph{Proceedings of the IEEE/CVF Conference on Computer Vision and Pattern Recognition}, pp.\  6712--6722, 2024.

\bibitem[Geng et~al.(2025)Geng, Herrmann, Hur, Cole, Sun, Wang, Pfaff, Lopez-Guevara, Doersch, Aytar, Rubinstein, Owens, and Sun]{geng2025motion}
Daniel Geng, Charles Herrmann, Junhwa Hur, Forrester Cole, Chen Sun, Oliver Wang, Tobias Pfaff, Tatiana Lopez-Guevara, Carl Doersch, Yusuf Aytar, Michael Rubinstein, Andrew Owens, and Deqing Sun.
\newblock Motion prompting: Controlling video generation with motion trajectories.
\newblock \emph{Computer Vision and Pattern Recognition (CVPR)}, 2025.

\bibitem[Geyer et~al.(2023)Geyer, Bar-Tal, Bagon, and Dekel]{tokenflow2023}
Michal Geyer, Omer Bar-Tal, Shai Bagon, and Tali Dekel.
\newblock Tokenflow: Consistent diffusion features for consistent video editing.
\newblock \emph{arXiv preprint arxiv:2307.10373}, 2023.

\bibitem[Hao et~al.(2018)Hao, Huang, and Belongie]{hao2018controllable}
Zekun Hao, Xun Huang, and Serge Belongie.
\newblock Controllable video generation with sparse trajectories.
\newblock In \emph{Proceedings of the IEEE Conference on Computer Vision and Pattern Recognition}, pp.\  7854--7863, 2018.

\bibitem[Harley et~al.(2022)Harley, Fang, and Fragkiadaki]{harley2022particle}
Adam~W Harley, Zhaoyuan Fang, and Katerina Fragkiadaki.
\newblock Particle video revisited: Tracking through occlusions using point trajectories.
\newblock In \emph{European Conference on Computer Vision}, pp.\  59--75. Springer, 2022.

\bibitem[He et~al.(2020)He, Fan, Wu, Xie, and Girshick]{he2020momentum}
Kaiming He, Haoqi Fan, Yuxin Wu, Saining Xie, and Ross Girshick.
\newblock Momentum contrast for unsupervised visual representation learning.
\newblock In \emph{Proceedings of the IEEE/CVF conference on computer vision and pattern recognition}, pp.\  9729--9738, 2020.

\bibitem[Ho \& Salimans(2022)Ho and Salimans]{ho2022classifier}
Jonathan Ho and Tim Salimans.
\newblock Classifier-free diffusion guidance.
\newblock \emph{arXiv preprint arXiv:2207.12598}, 2022.

\bibitem[Ho et~al.(2020)Ho, Jain, and Abbeel]{ho2020denoising}
Jonathan Ho, Ajay Jain, and Pieter Abbeel.
\newblock Denoising diffusion probabilistic models.
\newblock \emph{Advances in neural information processing systems}, 33:\penalty0 6840--6851, 2020.

\bibitem[Huang et~al.(2023)Huang, Herrmann, Hur, Lu, Sargent, Stone, Yang, and Sun]{huang2023self}
Hsin-Ping Huang, Charles Herrmann, Junhwa Hur, Erika Lu, Kyle Sargent, Austin Stone, Ming-Hsuan Yang, and Deqing Sun.
\newblock Self-supervised autoflow.
\newblock In \emph{Proceedings of the IEEE/CVF Conference on Computer Vision and Pattern Recognition}, pp.\  11412--11421, 2023.

\bibitem[Huang et~al.(2024)Huang, Zhang, Xu, He, Yu, Dong, Ma, Chanpaisit, Si, Jiang, Wang, Chen, Chen, Wang, Lin, Qiao, and Liu]{huang2024vbench++}
Ziqi Huang, Fan Zhang, Xiaojie Xu, Yinan He, Jiashuo Yu, Ziyue Dong, Qianli Ma, Nattapol Chanpaisit, Chenyang Si, Yuming Jiang, Yaohui Wang, Xinyuan Chen, Ying-Cong Chen, Limin Wang, Dahua Lin, Yu~Qiao, and Ziwei Liu.
\newblock {VBench++}: Comprehensive and versatile benchmark suite for video generative models.
\newblock \emph{arXiv preprint arXiv:2411.13503}, 2024.

\bibitem[Jabri et~al.(2020)Jabri, Owens, and Efros]{jabri2020space}
Allan Jabri, Andrew Owens, and Alexei Efros.
\newblock Space-time correspondence as a contrastive random walk.
\newblock \emph{Advances in neural information processing systems}, 33:\penalty0 19545--19560, 2020.

\bibitem[Jo \& Park(2019)Jo and Park]{jo2019sc}
Youngjoo Jo and Jongyoul Park.
\newblock Sc-fegan: Face editing generative adversarial network with user's sketch and color.
\newblock In \emph{Proceedings of the IEEE/CVF international conference on computer vision}, pp.\  1745--1753, 2019.

\bibitem[Jonschkowski et~al.(2020)Jonschkowski, Stone, Barron, Gordon, Konolige, and Angelova]{jonschkowski2020matters}
Rico Jonschkowski, Austin Stone, Jonathan~T Barron, Ariel Gordon, Kurt Konolige, and Anelia Angelova.
\newblock What matters in unsupervised optical flow.
\newblock In \emph{Computer Vision--ECCV 2020: 16th European Conference, Glasgow, UK, August 23--28, 2020, Proceedings, Part II 16}, 2020.

\bibitem[Karaev et~al.(2024{\natexlab{a}})Karaev, Makarov, Wang, Neverova, Vedaldi, and Rupprecht]{karaev2024cotracker3}
Nikita Karaev, Iurii Makarov, Jianyuan Wang, Natalia Neverova, Andrea Vedaldi, and Christian Rupprecht.
\newblock Cotracker3: Simpler and better point tracking by pseudo-labelling real videos.
\newblock \emph{arXiv preprint arXiv:2410.11831}, 2024{\natexlab{a}}.

\bibitem[Karaev et~al.(2024{\natexlab{b}})Karaev, Makarov, Wang, Neverova, Vedaldi, and Rupprecht]{karaev24cotracker3}
Nikita Karaev, Iurii Makarov, Jianyuan Wang, Natalia Neverova, Andrea Vedaldi, and Christian Rupprecht.
\newblock Cotracker3: Simpler and better point tracking by pseudo-labelling real videos.
\newblock In \emph{Proc. {arXiv:2410.11831}}, 2024{\natexlab{b}}.

\bibitem[Karaev et~al.(2024{\natexlab{c}})Karaev, Rocco, Graham, Neverova, Vedaldi, and Rupprecht]{karaev2024cotracker}
Nikita Karaev, Ignacio Rocco, Benjamin Graham, Natalia Neverova, Andrea Vedaldi, and Christian Rupprecht.
\newblock Cotracker: It is better to track together.
\newblock In \emph{European Conference on Computer Vision}, pp.\  18--35. Springer, 2024{\natexlab{c}}.

\bibitem[Karaev et~al.(2024{\natexlab{d}})Karaev, Rocco, Graham, Neverova, Vedaldi, and Rupprecht]{karaev23cotracker}
Nikita Karaev, Ignacio Rocco, Benjamin Graham, Natalia Neverova, Andrea Vedaldi, and Christian Rupprecht.
\newblock Cotracker: It is better to track together.
\newblock In \emph{Proc. {ECCV}}, 2024{\natexlab{d}}.

\bibitem[Karras et~al.(2024)Karras, Aittala, Kynk{\"a}{\"a}nniemi, Lehtinen, Aila, and Laine]{karras2024guiding}
Tero Karras, Miika Aittala, Tuomas Kynk{\"a}{\"a}nniemi, Jaakko Lehtinen, Timo Aila, and Samuli Laine.
\newblock Guiding a diffusion model with a bad version of itself.
\newblock \emph{Advances in Neural Information Processing Systems}, 37:\penalty0 52996--53021, 2024.

\bibitem[Kirillov et~al.(2023)Kirillov, Mintun, Ravi, Mao, Rolland, Gustafson, Xiao, Whitehead, Berg, Lo, et~al.]{kirillov2023segment}
Alexander Kirillov, Eric Mintun, Nikhila Ravi, Hanzi Mao, Chloe Rolland, Laura Gustafson, Tete Xiao, Spencer Whitehead, Alexander~C Berg, Wan-Yen Lo, et~al.
\newblock Segment anything.
\newblock In \emph{Proceedings of the IEEE/CVF international conference on computer vision}, pp.\  4015--4026, 2023.

\bibitem[Kojima et~al.(2022)Kojima, Gu, Reid, Matsuo, and Iwasawa]{kojima2022large}
Takeshi Kojima, Shixiang~Shane Gu, Machel Reid, Yutaka Matsuo, and Yusuke Iwasawa.
\newblock Large language models are zero-shot reasoners.
\newblock \emph{Advances in neural information processing systems}, 35:\penalty0 22199--22213, 2022.

\bibitem[Lai et~al.(2018)Lai, Huang, Wang, Shechtman, Yumer, and Yang]{Lai-ECCV-2018}
Wei-Sheng Lai, Jia-Bin Huang, Oliver Wang, Eli Shechtman, Ersin Yumer, and Ming-Hsuan Yang.
\newblock Learning blind video temporal consistency.
\newblock In \emph{European Conference on Computer Vision}, 2018.

\bibitem[Li et~al.(2023)Li, Zhang, Xu, Liu, Zhang, Ni, and Shum]{li2023mask}
Feng Li, Hao Zhang, Huaizhe Xu, Shilong Liu, Lei Zhang, Lionel~M Ni, and Heung-Yeung Shum.
\newblock Mask dino: Towards a unified transformer-based framework for object detection and segmentation.
\newblock In \emph{Proceedings of the IEEE/CVF conference on computer vision and pattern recognition}, pp.\  3041--3050, 2023.

\bibitem[Lipman et~al.(2022)Lipman, Chen, Ben-Hamu, Nickel, and Le]{Lipman2022FlowMF}
Yaron Lipman, Ricky T.~Q. Chen, Heli Ben-Hamu, Maximilian Nickel, and Matt Le.
\newblock Flow matching for generative modeling.
\newblock \emph{ArXiv}, abs/2210.02747, 2022.
\newblock URL \url{https://api.semanticscholar.org/CorpusID:252734897}.

\bibitem[Liu et~al.(2021)Liu, Wan, Huang, Song, Han, Liao, Jiang, and Liu]{liu2021deflocnet}
Hongyu Liu, Ziyu Wan, Wei Huang, Yibing Song, Xintong Han, Jing Liao, Bin Jiang, and Wei Liu.
\newblock Deflocnet: Deep image editing via flexible low-level controls.
\newblock In \emph{Proceedings of the IEEE/CVF Conference on Computer Vision and Pattern Recognition}, pp.\  10765--10774, 2021.

\bibitem[Liu et~al.(2019)Liu, Lyu, King, and Xu]{liu2019selflow}
Pengpeng Liu, Michael Lyu, Irwin King, and Jia Xu.
\newblock Selflow: Self-supervised learning of optical flow.
\newblock In \emph{Proceedings of the IEEE/CVF conference on computer vision and pattern recognition}, pp.\  4571--4580, 2019.

\bibitem[Liu et~al.(2024)Liu, Zeng, Ren, Li, Zhang, Yang, Jiang, Li, Yang, Su, et~al.]{liu2024grounding}
Shilong Liu, Zhaoyang Zeng, Tianhe Ren, Feng Li, Hao Zhang, Jie Yang, Qing Jiang, Chunyuan Li, Jianwei Yang, Hang Su, et~al.
\newblock Grounding dino: Marrying dino with grounded pre-training for open-set object detection.
\newblock In \emph{European Conference on Computer Vision}, pp.\  38--55. Springer, 2024.

\bibitem[Lugmayr et~al.(2022)Lugmayr, Danelljan, Romero, Yu, Timofte, and Van~Gool]{lugmayr2022repaint}
Andreas Lugmayr, Martin Danelljan, Andres Romero, Fisher Yu, Radu Timofte, and Luc Van~Gool.
\newblock Repaint: Inpainting using denoising diffusion probabilistic models.
\newblock In \emph{Proceedings of the IEEE/CVF conference on computer vision and pattern recognition}, pp.\  11461--11471, 2022.

\bibitem[Luo et~al.(2023)Luo, Dunlap, Park, Holynski, and Darrell]{luo2023diffusion}
Grace Luo, Lisa Dunlap, Dong~Huk Park, Aleksander Holynski, and Trevor Darrell.
\newblock Diffusion hyperfeatures: Searching through time and space for semantic correspondence.
\newblock \emph{Advances in Neural Information Processing Systems}, 36:\penalty0 47500--47510, 2023.

\bibitem[Meng et~al.(2021)Meng, He, Song, Song, Wu, Zhu, and Ermon]{meng2021sdedit}
Chenlin Meng, Yutong He, Yang Song, Jiaming Song, Jiajun Wu, Jun-Yan Zhu, and Stefano Ermon.
\newblock Sdedit: Guided image synthesis and editing with stochastic differential equations.
\newblock \emph{arXiv preprint arXiv:2108.01073}, 2021.

\bibitem[Nam et~al.(2025)Nam, Son, Chung, Kim, Jin, Hur, and Kim]{nam2025emergent}
Jisu Nam, Soowon Son, Dahyun Chung, Jiyoung Kim, Siyoon Jin, Junhwa Hur, and Seungryong Kim.
\newblock Emergent temporal correspondences from video diffusion transformers.
\newblock \emph{arXiv preprint arXiv:2506.17220}, 2025.

\bibitem[Neoral et~al.(2024)Neoral, {\v{S}}er{\`y}ch, and Matas]{neoral2024mft}
Michal Neoral, Jon{\'a}{\v{s}} {\v{S}}er{\`y}ch, and Ji{\v{r}}{\'\i} Matas.
\newblock Mft: Long-term tracking of every pixel.
\newblock In \emph{Proceedings of the IEEE/CVF Winter Conference on Applications of Computer Vision}, pp.\  6837--6847, 2024.

\bibitem[Nichol et~al.(2021)Nichol, Dhariwal, Ramesh, Shyam, Mishkin, McGrew, Sutskever, and Chen]{nichol2021glide}
Alex Nichol, Prafulla Dhariwal, Aditya Ramesh, Pranav Shyam, Pamela Mishkin, Bob McGrew, Ilya Sutskever, and Mark Chen.
\newblock Glide: Towards photorealistic image generation and editing with text-guided diffusion models.
\newblock \emph{arXiv preprint arXiv:2112.10741}, 2021.

\bibitem[Oquab et~al.(2023)Oquab, Darcet, Moutakanni, Vo, Szafraniec, Khalidov, Fernandez, Haziza, Massa, El-Nouby, et~al.]{oquab2023dinov2}
Maxime Oquab, Timoth{\'e}e Darcet, Th{\'e}o Moutakanni, Huy Vo, Marc Szafraniec, Vasil Khalidov, Pierre Fernandez, Daniel Haziza, Francisco Massa, Alaaeldin El-Nouby, et~al.
\newblock Dinov2: Learning robust visual features without supervision.
\newblock \emph{arXiv preprint arXiv:2304.07193}, 2023.

\bibitem[Perazzi et~al.(2016)Perazzi, Pont-Tuset, McWilliams, Van~Gool, Gross, and Sorkine-Hornung]{perazzi2016benchmark}
Federico Perazzi, Jordi Pont-Tuset, Brian McWilliams, Luc Van~Gool, Markus Gross, and Alexander Sorkine-Hornung.
\newblock A benchmark dataset and evaluation methodology for video object segmentation.
\newblock In \emph{Proceedings of the IEEE conference on computer vision and pattern recognition}, pp.\  724--732, 2016.

\bibitem[Podell et~al.(2023)Podell, English, Lacey, Blattmann, Dockhorn, M{\"u}ller, Penna, and Rombach]{podell2023sdxl}
Dustin Podell, Zion English, Kyle Lacey, Andreas Blattmann, Tim Dockhorn, Jonas M{\"u}ller, Joe Penna, and Robin Rombach.
\newblock Sdxl: Improving latent diffusion models for high-resolution image synthesis.
\newblock \emph{arXiv preprint arXiv:2307.01952}, 2023.

\bibitem[Polyak et~al.(2024)Polyak, Zohar, Brown, Tjandra, Sinha, Lee, Vyas, Shi, Ma, Chuang, Yan, Choudhary, Wang, Sethi, Pang, Ma, Misra, Hou, Wang, ran Jagadeesh, Li, Zhang, Singh, Williamson, Le, Yu, Singh, Zhang, Vajda, Duval, Girdhar, Sumbaly, Rambhatla, Tsai, Azadi, Datta, Chen, Bell, Ramaswamy, Sheynin, Bhattacharya, Motwani, Xu, Li, Hou, Hsu, Yin, Dai, Taigman, Luo, Liu, Wu, Zhao, Kirstain, He, He, Pumarola, Thabet, Sanakoyeu, Mallya, Guo, Araya, Kerr, Wood, Liu, Peng, Vengertsev, Schonfeld, Blanchard, Juefei-Xu, Nord, Liang, Hoffman, Kohler, Fire, Sivakumar, Chen, Yu, Gao, Georgopoulos, Moritz, Sampson, Li, Parmeggiani, Fine, Fowler, Petrovic, and Du]{Polyak2024MovieGA}
Adam Polyak, Amit Zohar, Andrew Brown, Andros Tjandra, Animesh Sinha, Ann Lee, Apoorv Vyas, Bowen Shi, Chih-Yao Ma, Ching-Yao Chuang, David Yan, Dhruv Choudhary, Dingkang Wang, Geet Sethi, Guan Pang, Haoyu Ma, Ishan Misra, Ji~Hou, Jialiang Wang, Ki~ran Jagadeesh, Kunpeng Li, Luxin Zhang, Mannat Singh, Mary Williamson, Matt Le, Matthew Yu, Mitesh~Kumar Singh, Peizhao Zhang, Peter Vajda, Quentin Duval, Rohit Girdhar, Roshan Sumbaly, Sai~Saketh Rambhatla, Sam~S. Tsai, Samaneh Azadi, Samyak Datta, Sanyuan Chen, Sean Bell, Sharadh Ramaswamy, Shelly Sheynin, Siddharth Bhattacharya, Simran Motwani, Tao Xu, Tianhe Li, Tingbo Hou, Wei-Ning Hsu, Xi~Yin, Xiaoliang Dai, Yaniv Taigman, Yaqiao Luo, Yen-Cheng Liu, Yi-Chiao Wu, Yue Zhao, Yuval Kirstain, Zecheng He, Zijian He, Albert Pumarola, Ali~K. Thabet, Artsiom Sanakoyeu, Arun Mallya, Baishan Guo, Boris Araya, Breena Kerr, Carleigh Wood, Ce~Liu, Cen Peng, Dimitry Vengertsev, Edgar Schonfeld, Elliot Blanchard, Felix Juefei-Xu, Fraylie Nord, Jeff Liang, John Hoffman, Jonas
  Kohler, Kaolin Fire, Karthik Sivakumar, Lawrence Chen, Licheng Yu, Luya Gao, Markos Georgopoulos, Rashel Moritz, Sara~K. Sampson, Shikai Li, Simone Parmeggiani, Steve Fine, Tara Fowler, Vladan Petrovic, and Yuming Du.
\newblock Movie gen: A cast of media foundation models.
\newblock \emph{ArXiv}, abs/2410.13720, 2024.
\newblock URL \url{https://api.semanticscholar.org/CorpusID:273403698}.

\bibitem[Radford et~al.(2021)Radford, Kim, Hallacy, Ramesh, Goh, Agarwal, Sastry, Askell, Mishkin, Clark, et~al.]{radford2021learning}
Alec Radford, Jong~Wook Kim, Chris Hallacy, Aditya Ramesh, Gabriel Goh, Sandhini Agarwal, Girish Sastry, Amanda Askell, Pamela Mishkin, Jack Clark, et~al.
\newblock Learning transferable visual models from natural language supervision.
\newblock In \emph{International conference on machine learning}, pp.\  8748--8763. PmLR, 2021.

\bibitem[Raffel et~al.(2020)Raffel, Shazeer, Roberts, Lee, Narang, Matena, Zhou, Li, and Liu]{raffel2020exploring}
Colin Raffel, Noam Shazeer, Adam Roberts, Katherine Lee, Sharan Narang, Michael Matena, Yanqi Zhou, Wei Li, and Peter~J Liu.
\newblock Exploring the limits of transfer learning with a unified text-to-text transformer.
\newblock \emph{Journal of machine learning research}, 21\penalty0 (140):\penalty0 1--67, 2020.

\bibitem[Rombach et~al.(2022)Rombach, Blattmann, Lorenz, Esser, and Ommer]{rombach2022high}
Robin Rombach, Andreas Blattmann, Dominik Lorenz, Patrick Esser, and Bj{\"o}rn Ommer.
\newblock High-resolution image synthesis with latent diffusion models.
\newblock In \emph{Proceedings of the IEEE/CVF conference on computer vision and pattern recognition}, pp.\  10684--10695, 2022.

\bibitem[Ruiz et~al.(2023)Ruiz, Li, Jampani, Pritch, Rubinstein, and Aberman]{ruiz2023dreambooth}
Nataniel Ruiz, Yuanzhen Li, Varun Jampani, Yael Pritch, Michael Rubinstein, and Kfir Aberman.
\newblock Dreambooth: Fine tuning text-to-image diffusion models for subject-driven generation.
\newblock In \emph{Proceedings of the IEEE/CVF conference on computer vision and pattern recognition}, pp.\  22500--22510, 2023.

\bibitem[Sand \& Teller(2008)Sand and Teller]{sand2008particle}
Peter Sand and Seth Teller.
\newblock Particle video: Long-range motion estimation using point trajectories.
\newblock \emph{International journal of computer vision}, 80:\penalty0 72--91, 2008.

\bibitem[Shrivastava \& Owens(2024)Shrivastava and Owens]{shrivastava2024gmrw}
Ayush Shrivastava and Andrew Owens.
\newblock Self-supervised any-point tracking by contrastive random walks.
\newblock In \emph{European Conference on Computer Vision (ECCV)}, 2024.
\newblock URL \url{https://arxiv.org/abs/2409.16288}.

\bibitem[Shtedritski et~al.(2023)Shtedritski, Rupprecht, and Vedaldi]{shtedritski2023does}
Aleksandar Shtedritski, Christian Rupprecht, and Andrea Vedaldi.
\newblock What does clip know about a red circle? visual prompt engineering for vlms.
\newblock In \emph{Proceedings of the IEEE/CVF International Conference on Computer Vision}, pp.\  11987--11997, 2023.

\bibitem[Si et~al.(2024)Si, Huang, Jiang, and Liu]{si2024freeu}
Chenyang Si, Ziqi Huang, Yuming Jiang, and Ziwei Liu.
\newblock Freeu: Free lunch in diffusion u-net.
\newblock In \emph{Proceedings of the IEEE/CVF Conference on Computer Vision and Pattern Recognition}, pp.\  4733--4743, 2024.

\bibitem[Sohl-Dickstein et~al.(2015)Sohl-Dickstein, Weiss, Maheswaranathan, and Ganguli]{sohl2015deep}
Jascha Sohl-Dickstein, Eric Weiss, Niru Maheswaranathan, and Surya Ganguli.
\newblock Deep unsupervised learning using nonequilibrium thermodynamics.
\newblock In \emph{International conference on machine learning}, pp.\  2256--2265. pmlr, 2015.

\bibitem[Stojanov et~al.(2025)Stojanov, Wendt, Kim, Venkatesh, Feigelis, Wu, and Yamins]{stojanov2025self}
Stefan Stojanov, David Wendt, Seungwoo Kim, Rahul Venkatesh, Kevin Feigelis, Jiajun Wu, and Daniel~LK Yamins.
\newblock Self-supervised learning of motion concepts by optimizing counterfactuals.
\newblock \emph{arXiv preprint arXiv:2503.19953}, 2025.

\bibitem[Sun et~al.(2018)Sun, Yang, Liu, and Kautz]{sun2018pwc}
Deqing Sun, Xiaodong Yang, Ming-Yu Liu, and Jan Kautz.
\newblock Pwc-net: Cnns for optical flow using pyramid, warping, and cost volume.
\newblock In \emph{Proceedings of the IEEE conference on computer vision and pattern recognition}, pp.\  8934--8943, 2018.

\bibitem[Tang et~al.(2023)Tang, Jia, Wang, Phoo, and Hariharan]{tang2023emergent}
Luming Tang, Menglin Jia, Qianqian Wang, Cheng~Perng Phoo, and Bharath Hariharan.
\newblock Emergent correspondence from image diffusion.
\newblock \emph{Advances in Neural Information Processing Systems}, 36:\penalty0 1363--1389, 2023.

\bibitem[Teed \& Deng(2020)Teed and Deng]{teed2020raft}
Zachary Teed and Jia Deng.
\newblock Raft: Recurrent all-pairs field transforms for optical flow.
\newblock In \emph{Computer Vision--ECCV 2020: 16th European Conference, Glasgow, UK, August 23--28, 2020, Proceedings, Part II 16}, pp.\  402--419. Springer, 2020.

\bibitem[Tong et~al.(2024)Tong, Brown, Wu, Woo, IYER, Akula, Yang, Yang, Middepogu, Wang, et~al.]{tong2024cambrian}
Peter Tong, Ellis Brown, Penghao Wu, Sanghyun Woo, Adithya Jairam~Vedagiri IYER, Sai~Charitha Akula, Shusheng Yang, Jihan Yang, Manoj Middepogu, Ziteng Wang, et~al.
\newblock Cambrian-1: A fully open, vision-centric exploration of multimodal llms.
\newblock \emph{Advances in Neural Information Processing Systems}, 37:\penalty0 87310--87356, 2024.

\bibitem[Venkatesh et~al.(2023)Venkatesh, Chen, Feigelis, Bear, Jedoui, Kotar, Binder, Lee, Liu, Smith, et~al.]{venkatesh2023understanding}
Rahul Venkatesh, Honglin Chen, Kevin Feigelis, Daniel~M Bear, Khaled Jedoui, Klemen Kotar, Felix Binder, Wanhee Lee, Sherry Liu, Kevin~A Smith, et~al.
\newblock Understanding physical dynamics with counterfactual world modeling.
\newblock \emph{arXiv preprint arXiv:2312.06721}, 2023.

\bibitem[Vondrick et~al.(2018)Vondrick, Shrivastava, Fathi, Guadarrama, and Murphy]{Vondrick_2018_ECCV}
Carl Vondrick, Abhinav Shrivastava, Alireza Fathi, Sergio Guadarrama, and Kevin Murphy.
\newblock Tracking emerges by colorizing videos.
\newblock In \emph{Proceedings of the European Conference on Computer Vision (ECCV)}, September 2018.

\bibitem[Wang et~al.(2025)Wang, Ai, Wen, Mao, Xie, Chen, Yu, Zhao, Yang, Zeng, et~al.]{wang2025wan}
Ang Wang, Baole Ai, Bin Wen, Chaojie Mao, Chen-Wei Xie, Di~Chen, Feiwu Yu, Haiming Zhao, Jianxiao Yang, Jianyuan Zeng, et~al.
\newblock Wan: Open and advanced large-scale video generative models.
\newblock \emph{arXiv preprint arXiv:2503.20314}, 2025.

\bibitem[Wang et~al.(2019)Wang, Jabri, and Efros]{wang2019learning}
Xiaolong Wang, Allan Jabri, and Alexei~A Efros.
\newblock Learning correspondence from the cycle-consistency of time.
\newblock In \emph{Proceedings of the IEEE/CVF conference on computer vision and pattern recognition}, 2019.

\bibitem[Wang et~al.(2023)Wang, Wang, Cao, Shen, and Huang]{wang2023images}
Xinlong Wang, Wen Wang, Yue Cao, Chunhua Shen, and Tiejun Huang.
\newblock Images speak in images: A generalist painter for in-context visual learning.
\newblock In \emph{Proceedings of the IEEE/CVF Conference on Computer Vision and Pattern Recognition}, pp.\  6830--6839, 2023.

\bibitem[Wei et~al.(2022)Wei, Wang, Schuurmans, Bosma, Xia, Chi, Le, Zhou, et~al.]{wei2022chain}
Jason Wei, Xuezhi Wang, Dale Schuurmans, Maarten Bosma, Fei Xia, Ed~Chi, Quoc~V Le, Denny Zhou, et~al.
\newblock Chain-of-thought prompting elicits reasoning in large language models.
\newblock \emph{Advances in neural information processing systems}, 35:\penalty0 24824--24837, 2022.

\bibitem[Wu et~al.(2024)Wu, Si, Jiang, Huang, and Liu]{wu2024freeinit}
Tianxing Wu, Chenyang Si, Yuming Jiang, Ziqi Huang, and Ziwei Liu.
\newblock Freeinit: Bridging initialization gap in video diffusion models.
\newblock In \emph{European Conference on Computer Vision}, pp.\  378--394. Springer, 2024.

\bibitem[Yang et~al.(2024{\natexlab{a}})Yang, Kang, Huang, Xu, Feng, and Zhao]{yang2024depth}
Lihe Yang, Bingyi Kang, Zilong Huang, Xiaogang Xu, Jiashi Feng, and Hengshuang Zhao.
\newblock Depth anything: Unleashing the power of large-scale unlabeled data.
\newblock In \emph{Proceedings of the IEEE/CVF Conference on Computer Vision and Pattern Recognition}, pp.\  10371--10381, 2024{\natexlab{a}}.

\bibitem[Yang et~al.(2024{\natexlab{b}})Yang, Teng, Zheng, Ding, Huang, Xu, Yang, Hong, Zhang, Feng, Yin, Gu, Zhang, Wang, Cheng, Liu, Xu, Dong, and Tang]{Yang2024CogVideoXTD}
Zhuoyi Yang, Jiayan Teng, Wendi Zheng, Ming Ding, Shiyu Huang, Jiazheng Xu, Yuanming Yang, Wenyi Hong, Xiaohan Zhang, Guanyu Feng, Da~Yin, Xiaotao Gu, Yuxuan Zhang, Weihan Wang, Yean Cheng, Ting Liu, Bin Xu, Yuxiao Dong, and Jie Tang.
\newblock Cogvideox: Text-to-video diffusion models with an expert transformer.
\newblock \emph{ArXiv}, abs/2408.06072, 2024{\natexlab{b}}.
\newblock URL \url{https://api.semanticscholar.org/CorpusID:271855655}.

\bibitem[Yao et~al.(2024)Yao, Zhang, Zhang, Liu, Chua, and Sun]{yao2024cpt}
Yuan Yao, Ao~Zhang, Zhengyan Zhang, Zhiyuan Liu, Tat-Seng Chua, and Maosong Sun.
\newblock Cpt: Colorful prompt tuning for pre-trained vision-language models.
\newblock \emph{AI Open}, 5:\penalty0 30--38, 2024.

\bibitem[Yu et~al.(2023)Yu, Sohn, Kim, and Shin]{Yu2023VideoPD}
Sihyun Yu, Kihyuk Sohn, Subin Kim, and Jinwoo Shin.
\newblock Video probabilistic diffusion models in projected latent space.
\newblock \emph{2023 IEEE/CVF Conference on Computer Vision and Pattern Recognition (CVPR)}, 2023.

\bibitem[Zhai et~al.(2023)Zhai, Mustafa, Kolesnikov, and Beyer]{zhai2023sigmoid}
Xiaohua Zhai, Basil Mustafa, Alexander Kolesnikov, and Lucas Beyer.
\newblock Sigmoid loss for language image pre-training.
\newblock In \emph{Proceedings of the IEEE/CVF international conference on computer vision}, pp.\  11975--11986, 2023.

\bibitem[Zhang et~al.(2023{\natexlab{a}})Zhang, Herrmann, Hur, Polania~Cabrera, Jampani, Sun, and Yang]{zhang2023tale}
Junyi Zhang, Charles Herrmann, Junhwa Hur, Luisa Polania~Cabrera, Varun Jampani, Deqing Sun, and Ming-Hsuan Yang.
\newblock A tale of two features: Stable diffusion complements dino for zero-shot semantic correspondence.
\newblock \emph{Advances in Neural Information Processing Systems}, 36:\penalty0 45533--45547, 2023{\natexlab{a}}.

\bibitem[Zhang et~al.(2023{\natexlab{b}})Zhang, Rao, and Agrawala]{zhang2023adding}
Lvmin Zhang, Anyi Rao, and Maneesh Agrawala.
\newblock Adding conditional control to text-to-image diffusion models.
\newblock In \emph{Proceedings of the IEEE/CVF international conference on computer vision}, pp.\  3836--3847, 2023{\natexlab{b}}.

\bibitem[Zhang et~al.(2016)Zhang, Isola, and Efros]{zhang2016colorful}
Richard Zhang, Phillip Isola, and Alexei~A Efros.
\newblock Colorful image colorization.
\newblock In \emph{Computer Vision--ECCV 2016: 14th European Conference, Amsterdam, The Netherlands, October 11-14, 2016, Proceedings, Part III 14}, pp.\  649--666. Springer, 2016.

\bibitem[Zhang et~al.(2023{\natexlab{c}})Zhang, Wei, Jiang, Zhang, Zuo, and Tian]{zhang2023controlvideo}
Yabo Zhang, Yuxiang Wei, Dongsheng Jiang, Xiaopeng Zhang, Wangmeng Zuo, and Qi~Tian.
\newblock Controlvideo: Training-free controllable text-to-video generation.
\newblock \emph{arXiv preprint arXiv:2305.13077}, 2023{\natexlab{c}}.

\bibitem[Zheng et~al.(2023)Zheng, Harley, Shen, Wetzstein, and Guibas]{zheng2023pointodyssey}
Yang Zheng, Adam~W Harley, Bokui Shen, Gordon Wetzstein, and Leonidas~J Guibas.
\newblock Pointodyssey: A large-scale synthetic dataset for long-term point tracking.
\newblock In \emph{Proceedings of the IEEE/CVF International Conference on Computer Vision}, pp.\  19855--19865, 2023.

\bibitem[Zholus et~al.(2025)Zholus, Doersch, Yang, Koppula, Patraucean, He, Rocco, Sajjadi, Chandar, and Goroshin]{zholus2025tapnext}
Artem Zholus, Carl Doersch, Yi~Yang, Skanda Koppula, Viorica Patraucean, Xu~Owen He, Ignacio Rocco, Mehdi S.~M. Sajjadi, Sarath Chandar, and Ross Goroshin.
\newblock Tapnext: Tracking any point (tap) as next token prediction.
\newblock \emph{arXiv preprint arXiv:2504.05579}, 2025.

\bibitem[Zhou et~al.(2022)Zhou, Yang, Loy, and Liu]{zhou2022learning}
Kaiyang Zhou, Jingkang Yang, Chen~Change Loy, and Ziwei Liu.
\newblock Learning to prompt for vision-language models.
\newblock \emph{International Journal of Computer Vision}, 130\penalty0 (9):\penalty0 2337--2348, 2022.

\bibitem[Zhou et~al.(2024)Zhou, Yang, Wang, Luo, and Loy]{zhou2024upscale}
Shangchen Zhou, Peiqing Yang, Jianyi Wang, Yihang Luo, and Chen~Change Loy.
\newblock Upscale-a-video: Temporal-consistent diffusion model for real-world video super-resolution.
\newblock In \emph{Proceedings of the IEEE/CVF Conference on Computer Vision and Pattern Recognition}, pp.\  2535--2545, 2024.

\bibitem[Zhuang et~al.(2021)Zhuang, Koyejo, and Schwing]{zhuang2021enjoy}
Peiye Zhuang, Oluwasanmi Koyejo, and Alexander~G Schwing.
\newblock Enjoy your editing: Controllable gans for image editing via latent space navigation.
\newblock \emph{arXiv preprint arXiv:2102.01187}, 2021.

\end{thebibliography}

\newpage
\appendix

\section{Quantitative Results on TAP-Vid}
Table~\ref{tab:kubric} presents results on TAP-Vid Kubric (using a subset of 30 videos) with our method based on the Wan2.1-14B model. Our approach outperforms zero-shot baselines, consistent with the results reported in Table 1 of the main paper.

\begin{figure*}[h]
    \centering
    \includegraphics[width=\textwidth]{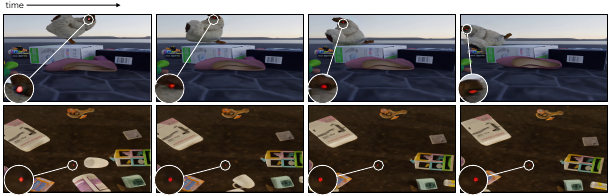}
    \vspace{-1.5mm}
    \vspace{-4mm}
    \caption{{\bf Qualitative Results on TAP-Vid Kubric.} The top row shows a successful example of point propagation. In contrast, the bottom row illustrates a failure case where the point is not propagated due to the surface having very low texture.}
    \label{fig:suppl_kubric}
\end{figure*}

However, the overall performance on Kubric is comparatively lower, likely due to the dataset’s synthetic nature. The scenes are generated using a graphics simulator and typically consist of simple environments with basic textures and objects exhibiting non-natural, erratic motion, as illustrated in Fig.~\ref{fig:suppl_kubric}. These characteristics introduce challenges for faithful video re-generation, which in turn impacts the accuracy of point propagation and tracking.

\begin{table*}[h!]

\centering

\begin{tabular}{ll ccc}
\toprule
\multirow{2}{*}{\textbf{Method}} &
\multirow{2}{*}{\textbf{Supervision}} &
\multicolumn{3}{c}{\textbf{TAP-Vid Kubric}}
\\

\cmidrule(lr){3-5}

& & AJ~$\uparrow$ & $<\delta^x_\textrm{avg}$~$\uparrow$ & OA~$\uparrow$ \\
\midrule
RAFT~\citep{teed2020raft}      & \multirow{5}{*}{Supervised} &
$68.50$ & $83.01$ & $89.94$ \\
TAP-Net~\citep{doersch2022tap}   &  & $68.22$ & $79.87$ & $93.35$ \\
TAPIR~\citep{doersch2023tapir}     &  & $87.88$ & $93.99$ & $96.09$ \\
CoTracker3~\citep{karaev24cotracker3}       &  & $76.99$ & $92.35$ & $92.35$ \\
TAPNext~\citep{zholus2025tapnext}           &  & $80.91$ & $87.03$ & $97.16$ \\

\midrule

GMRW~\citep{shrivastava2024gmrw}      & \multirow{2}{*}{Self-Sup.}        & $55.04$ & $72.22$ & $84.67$ \\
Opt-CWM~\citep{stojanov2025self}           &        & $60.11$ & $77.24$ & $85.62$ \\

\midrule

DINOv2+NN~\citep{oquab2023dinov2} &\multirow{5}{*}{Zero-Shot}  & $20.10$ & $40.25$ & $53.27$ \\
DIFT~\citep{tang2023emergent}   &   &   $25.93$ & $40.12$ & $74.08$ \\
SD-DINO~\citep{zhang2023tale}   &   & $28.89$ & $47.11$ & $47.10$ \\
Ours              &   & $31.51$ & $38.42$ & $53.23$ \\
Ours (upsampled)     &   & $33.55$ & $40.02$ & $54.80$ \\

\bottomrule
\end{tabular}

\caption{{\bf TAP-Vid Kubric Results.} We show results on TAP-Vid Kubric with \emph{first} sampling strategy.}

\label{tab:kubric}
\end{table*}

\subsection{Ablations}

\xhdr{Tracker Ablations.} We ablate key components of our tracking pipeline. First, we run the tracker without any enhancements on the generated videos. Adding a local search window around the previously detected point provides a small improvement, especially under occlusion. Gradually expanding the search radius when the query point becomes occluded yields further gains. We then introduce a position–refinement step that averages the coordinates of all red pixels within a fixed neighborhood around the predicted point, achieving the best overall performance. Finally, replacing the HSV color space with LAB causes a slight drop in accuracy, indicating that HSV is better suited for red-dot detection in our setup. Results are shown in Table \ref{tab:ablations-tracker}.

\begin{table*}[h]
\centering
\small
{\resizebox{0.9\linewidth}{!}{
\begin{tabular}{cccc ccc}

\toprule

\multirow{2}{*}{\textbf{Color}} &
\multirow{2}{*}{\textbf{Local search }} &
\multirow{2}{*}{\textbf{Occlusion based }} &
\multirow{2}{*}{\textbf{Average over}} &
\multicolumn{3}{c}{\textbf{TAP-Vid DAVIS}}
\\

\cmidrule(lr){5-7}

\textbf{space} &
\textbf{window} &
\textbf{search radius} &
\textbf{color pixels} &
AJ~$\uparrow$ & $<\delta^x_\textrm{avg}$~$\uparrow$ & OA~$\uparrow$ \\
\midrule
HSV &        &        &        & 35.80 & 53.15 & 81.79 \\
HSV & \cmark &        &        & 38.92 & 53.55 & 84.92 \\
HSV & \cmark & \cmark &        & 39.08 & 54.57 & 85.07 \\
\grayback{HSV} & \grayback{\cmark} & \grayback{\cmark} & \grayback{\cmark} & \grayback{42.70} & \grayback{59.26} & \grayback{85.14} \\
LAB & \cmark & \cmark & \cmark & 42.30 & 57.81 & 84.84 \\
\bottomrule
\end{tabular}
}}

\caption{{\bf Tracker Ablations.} (Sec.~\ref{sec:background}). We assess local search window, adaptive radius for occlusions, averaging red pixel positions, and performance across HSV vs. LAB color spaces.}
\label{tab:ablations-tracker}
\vspace{-2mm}
\end{table*}

\xhdr{Additional ablations.} We further assess model hyperparameters on a subset of TAP-Vid DAVIS videos (Table \ref{tab:ablations-cfg}). We ablate the parameter $\lambda$ (Eq. 5, main paper), which weights the noise estimate from the edited image. The best performance occurs at $\lambda = 8$. Table \ref{tab:marker-color-ablation} reports results when varying the marker color. While our approach is robust to different marker colors, using red provides a slight performance gain.

\vspace{-2mm}

\begin{figure}[h!]
  \centering
  \begin{minipage}[b]{0.6\linewidth}
    \centering
    \small
    \resizebox{\linewidth}{!}{

\setlength{\tabcolsep}{16pt}
\begin{tabular}{l ccc}
\toprule
\multirow{2}{*}{\textbf{Method}} &
\multicolumn{3}{c}{\textbf{TAP-Vid DAVIS}}
\\

\cmidrule(lr){2-4}
& AJ~$\uparrow$ & $<\delta^x_\textrm{avg}$~$\uparrow$ & OA~$\uparrow$ \\
\midrule

$\lambda=4$ & 34.60 & 52.48 & 77.94 \\
$\lambda=8$ & 35.54 & 52.98 & 78.80 \\
$\lambda=11$ & 32.82 & 52.08 & 75.66 \\
$\lambda=14$ & 31.92 & 52.13 & 74.09 \\

\bottomrule
\end{tabular}

    }
    \captionof{table}{{\bf Counterfactual Enhancement Guidance.} {We present ablation results for different values of $\lambda$, which controls the influence of the noise estimate from the edited image (with the colored dot) in counterfactual enhancement guidance.}}
    \label{tab:ablations-cfg}
  \end{minipage}\hfill
  \vspace{-5mm}
  \begin{minipage}[b]{0.35\linewidth}
    \centering
    \small
    \resizebox{\linewidth}{!}{
      \begin{tabular}{l ccc}
\toprule
\multirow{2}{*}{\textbf{Color}} &
\multicolumn{3}{c}{\textbf{TAP-Vid DAVIS}}
\\

\cmidrule(lr){2-4}
& AJ~$\uparrow$ & $<\delta^x_\textrm{avg}$~$\uparrow$ & OA~$\uparrow$ \\
\midrule

Red & 48.60 & 63.47 & 85.75 \\
Blue & 46.51 & 60.80 & 84.08 \\
\bottomrule
\end{tabular}

\label{tab:ablations-marker-color}

    }
    \captionof{table}{{\bf Marker color.} We use different marker colors as prompt to show that our approach is invariant to marker color.}
    \label{tab:marker-color-ablation}
  \end{minipage}
\end{figure}

\subsection{V-Bench scores}
Table~\ref{tab:vbench} shows tracking performance alongside VBench~\citep{huang2024vbench++} scores for Wan2.1 (1.3B and 14B variants), and CogVideoX~\citep{Yang2024CogVideoXTD}. VBench I2V benchmark evaluates the generation quality of image-conditioned video models. Tracking and generation quality both improve progressively from CogVideoX to Wan2.1-1.3B and further to Wan2.1-14B. We attribute this to the higher video generation quality—reflected in the superior VBench scores—which suggests that better generative models can directly boost tracking accuracy.
\begin{table}[h]
\small
\centering
\setlength{\tabcolsep}{2pt}
\begin{tabular}{l cc}
\toprule
\multirow{2}{*}{\textbf{Method}} &
\textbf{TAP-Vid DAVIS} &
\textbf{VBench}
\\

\cmidrule(lr){2-2}
\cmidrule(lr){3-3}

& AJ~$\uparrow$  &
Total Score
\\

\midrule

CogVideoX1.5-5B~\citep{Yang2024CogVideoXTD} & 24.15 & 71.58 \\

Wan2.1-1.3B~\citep{wang2025wan}         & 44.58 & 83.26 \\
Wan2.1-14B~\citep{wang2025wan}          & 48.60 & 86.66 \\

\bottomrule
\end{tabular}
\caption{{\bf VBench~\citep{huang2024vbench++} results.} We show VBench numbers for the different video models used.}
\label{tab:vbench}
\end{table}

\section{Implementation Details}

\subsection{Video Preprocessing}
\label{sec:video-preprocess}

\xhdr{Color Rebalancing.}
Our tracker identifies red pixels in each frame as predicted points. To avoid false positives, we first remove red pixels from the original frame. We convert the frame to the HSV color space and detect pixels whose hue values fall within $[-30^\circ, 10^\circ]$, and whose saturation and value lie inside an ellipse with semi-major and semi-minor axes $r_1 = 80$, $r_2 = 30$, centered at $(255, 255)$. For detected red pixels, we clip the saturation to a maximum of 80, effectively desaturating them.

\xhdr{Padding Input Video.}
Both Wan and CogVideoX require that the input video contains $4T + 1$ frames. To satisfy this constraint, we pad the input by repeating the last frame until this condition is met. After re-generation, the added frames are removed to restore the original length.

\xhdr{Video Upscaling.}
We observe that using high-resolution videos improves point propagation, reducing generation artifacts and minimizing drift. To upscale the input videos, we use Upscale-A-Video~\citep{zhou2024upscale}, a diffusion-based video upscaling method. Starting from $256 \times 256$ input resolution (from TAP-Vid), we upscale to $1024 \times 1024$ using Upscale-A-Video, then downscale to $480 \times 832$ to match the video model’s expected resolution. For final tracking evaluation, we resize the output back to $256 \times 256$.

\subsection{Point Propagation}
\looseness=-1
As described in Sec.~3.1, we use SDEdit with a denoising strength $\gamma = 0.5$ to control the signal-to-noise ratio. The diffusion timestep $t$ is calculated based on $\gamma$ and the total number of diffusion steps $T$:
\begin{equation} \label{eq:timestep}
t = \floor*{\gamma \cdot T}
\end{equation}
\xhdr{Counterfactual Enhancement Guidance}
To enhance the effect of the guidance from the edited image (with a colored dot), we use Eq.~5 (main paper) to compute the noise estimate. In our experiments, we follow the traditional classifier-free guidance scheme, where the guidance weight $\lambda$ is set to 8.

\subsection{Tracker}

\subsubsection{Marker Detection}
To identify the marker in the generated image, we perform color thresholding in the HSV color space. Specifically, we define the hue range as (H-10, H+5), the saturation range as (150, 255), and the value range as (150, 255). A pixel is considered as a potential marker pixel if its HSV components fall within this interval.

\subsubsection{Local Search and Occlusion Handling}

To effectively locate the marker in each frame, we constrain our search for red pixels to a circular region of radius $r$ centered at the previous detection. By default, this search radius is set to $r_{\text{default}} = 90$. If an occlusion is detected in the previous frame, we expand the search region to accommodate the increased positional uncertainty:
\begin{equation}
r = \min(r_{\text{default}} \times 1.1, r_{\text{max}})
\end{equation}
where $r_{\text{max}} = 150$. Once the marker is successfully detected again, we reset $r$ to its default value to maintain efficiency and avoid spurious detections.

\subsubsection{Center Estimation}

After identifying candidate red pixels, we first select the one closest to the previous detection as an anchor. Around this anchor point, we examine a $20$-pixel radius to gather nearby red pixel detections. The final predicted tracking point for the current frame is computed as the average position of these collected pixels. This averaging process produces a stable and consistent estimate for the red blob’s center, leading to robust and accurate tracking across frames.

\end{document}